\documentclass[10pt,twocolumn,letterpaper]{article}

\usepackage{cvpr}
\usepackage{float}
\usepackage{times}
\usepackage{epsfig}
\usepackage{graphicx}
\usepackage{amsmath}
\usepackage{amssymb}
\usepackage{color}
\usepackage{enumerate}
\usepackage{multirow}
\usepackage{algpseudocode,algorithm,algorithmicx}
\makeatletter
\newcommand{\thickhline}{%
	\noalign {\ifnum 0=`}\fi \hrule height 1pt
	\futurelet \reserved@a \@xhline
}
\makeatother

\usepackage[breaklinks=true,bookmarks=false]{hyperref}

\cvprfinalcopy 

\def\cvprPaperID{1182} 

\begin{document}

\title{Blending-target Domain Adaptation by Adversarial Meta-Adaptation Networks}

\author{Ziliang Chen$^{1}$, \ Jingyu Zhuang$^{1}$, \ Xiaodan Liang$^{1,2}$, \ Liang Lin$^{1,2}$\thanks{Corresponding author: Liang Lin.}\\ $^1$Sun Yat-sen University  \ \ $^2$DarkMatter AI Research\\
	\tt\small c.ziliang@yahoo.com, \tt\small zhuangjy6@mail2.sysu.edu.cn, \tt\small xdliang328@gmail.com, \tt\small linliang@ieee.org 
}

\twocolumn[{%
	 \maketitle \begin{figure}[H]\vspace{-12pt} \hsize=\textwidth
	 	 \centering \includegraphics[width=17cm]{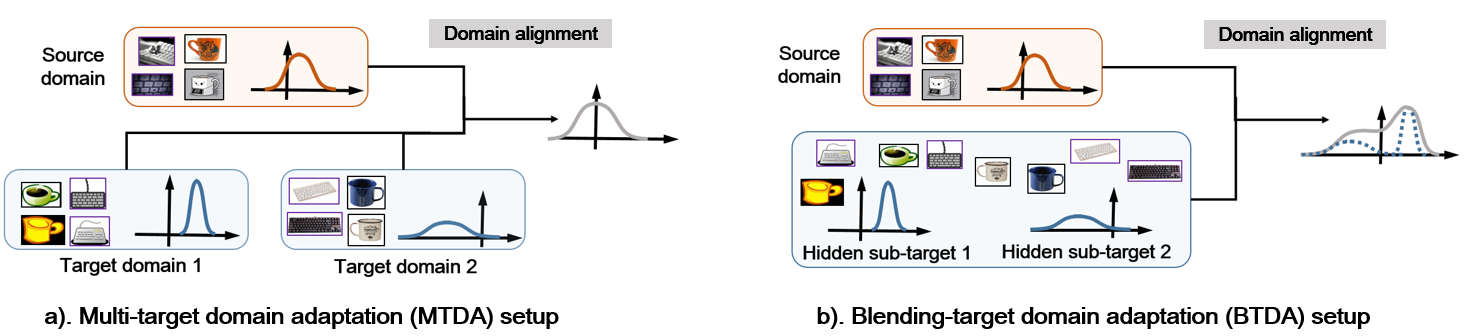} \vspace{6pt} \caption{The comparison of MTDA and BTDA (color orange and blue denote source and target) setups. In MTDA (a), target domains are explicitly separated and we are informed by which target an unlabeled sample originates from. In BTDA (b), sub-target IDs are encrypted. If we treat them as a combined single target, transfer learning will lead to the adaptation on a multi-target mixture instead of each hidden target (gray distribution curves in (a), (b)). It implies category-shifted adaptation and negative transfer in practice. Best viewed in color.}\label{mtda} \vspace{8pt}\end{figure} }] 

\begin{abstract}
   (Unsupervised) Domain Adaptation (\textbf{DA}) seeks for classifying target instances when solely provided with source labeled and target unlabeled examples for training. Learning domain-invariant features helps to achieve this goal, whereas it underpins unlabeled samples drawn from a single or multiple explicit target domains (Multi-target DA). In this paper, we consider a more realistic transfer scenario: our target domain is comprised of 
   multiple sub-targets implicitly blended with each other, so that learners could not identify which sub-target each unlabeled sample belongs to.
   This Blending-target Domain Adaptation (\textbf{BTDA}) scenario commonly appears in practice and threatens the validities of most existing DA algorithms, due to the presence of domain gaps and categorical misalignments among these hidden sub-targets. 
   
     
   To reap the transfer performance gains in this new scenario, we propose Adversarial Meta-Adaptation Network (AMEAN). AMEAN entails two adversarial transfer learning processes. The first is a conventional adversarial transfer to bridge our source and mixed target domains. To circumvent the intra-target category misalignment, the second process presents as ``learning to adapt'': It deploys an unsupervised meta-learner receiving target data and their ongoing feature-learning feedbacks, to discover target clusters as our ``meta-sub-target'' domains. These meta-sub-targets auto-design our meta-sub-target DA loss, which empirically eliminates the implicit category mismatching in our mixed target.  
   We evaluate AMEAN and a variety of DA algorithms in three benchmarks under the BTDA setup. Empirical results show that BTDA is a quite challenging transfer setup for most existing DA algorithms, yet AMEAN significantly outperforms these state-of-the-art baselines and effectively restrains the negative transfer effects in BTDA. 
   \end{abstract}

\vspace{-4pt}\section{Introduction}\footnote{* Code is available at {{\color{red} http://github.com/zjy526223908/BTDA }}.}
Despite achieving a growing number of successes, deep supervised learning algorithms remain restrictive in a variety of new application scenarios due to their vulnerabilities towards \emph{domain shifts} \cite{gretton2009covariate}: when evaluated on newly-emerged unlabeled target examples drawn from a distribution non-identical with the training source density, supervised learners inevitably perform inferior. In terms of visual data, this issue stems from diverse domain-specific appearance variabilities, \emph{e.g.} differences in background and camera poses, occlusions and volatile illumination conditions, \emph{etc}. Hence the variabilities are highly relevant in most machine vision implementations and obstacle their advancements. To address these problems, (unsupervised) Domain Adaptations (DAs) choose suitable statistical measures between source and target domains, \emph{e.g.}, maximum mean discrepancy (MMD) \cite{long2015learning} and adversarial-network distance \cite{Ganin2017Domain,tzeng2017adversarial}, to learn domain-invariant features in pursuit of consistent cross-domain model performances. The related studies increasingly attract a large amount of interests from the areas of domain-adaptive perception \cite{hoffman2016fcns,Chen2018Domain,bousmalis2016unsupervised}, autonomous steering \cite{You2017Virtual,Yang2018Real} and robotic vision \cite{bousmalis2018using,balloch2018unbiasing,rusu2016sim}. 

Most existing DA approaches indeed have evidenced impressive performances in a laboratory, whereas the domain shifts that frequently occur in reality, are far from being settled through these techniques. One explanation is the ideal target-domain premise these DA techniques start from. Particularly, DAs are conventionally established on a ``single-target'' preset, namely, all target examples are drawn from an identical distribution. Some recent researches focus on \emph{Multi-target DA} (MTDA) \cite{anonymous2019unsupervised,Yu2018Multi}, where target examples stem from multiple distributions, whereas we exactly know which target they belong to (See Fig.\ref{mtda}.(a)). 

In this paper, we argue these target-domain preconditions always taken for granted in most previous transfer learning literatures. After revisiting widespread presences, we discover that target unlabeled examples are often too diverse to well suit a ``single-target'' foresight. For instance, virtual-to-real researches \cite{bousmalis2018using,You2017Virtual} encourage robots and driver agents trained on a simulation platform to adaptively perform in the real-world environment. However, the target real-world environment includes extensive arrays of scenarios and continuously changes as time goes by. Another case is an encrypted dataset stored in a cloud server \cite{gilad2016cryptonets}, where the unlabeled examples are derived from multiple origins whereas due to a privacy protection, users have no access to identify these origins. These facts imply the existence of multiple sub-target domains while unlike standard MTDA, these sub-targets are blended with each other so that learners are not able to identify which sub-target each unlabeled example belongs to (See Fig.\ref{mtda}.(b)). This so-called \emph{blending-target domain adaptation} (\textbf{BTDA}) scenario regularly occurs in more other circumstances and during adaptation process, it commonly arouses notorious negative transfer effects \cite{pan2010survey} by two reasons: 
\begin{itemize}
	\vspace{-6pt}\item Hidden sub-target domains are organized as a mixture distribution, whereas without the knowledge of sub-target ID, it is quite difficult to align the category to reduce their mismatching across the sub-targets. 
	\vspace{-6pt}\item Regardless of the domain gaps among the hidden targets, existing DA approaches will suffer from the category misalignments among these sub-targets. 
\end{itemize}. Due to the blending-target preset, BTDA can not be solved by existing multi-target transfer learning methods.

To reap transfer performance gains and simultaneously prevent the negative transfer effects in BTDA scenario, we propose \emph{Adversarial Meta-Adaptation Network} (\textbf{AMEAN}) attempting to solve this problem under the context of visual recognition. AMEAN evolves from popular adversarial adaptation frameworks \cite{ganin2014unsupervised,hoffman2017simultaneous} and opts for minimizing the discrepancy between our source and mixed target domains. But distinguished from these existing pipelines, our AMEAN is inspired by meta-learning and AutoML \cite{andrychowicz2016learning,Xu2018AutoLoss}, which concurrently deploys an \emph{unsupervised meta-learner} to learn deep target cluster embeddings by receiving the mixed target data and their ongoing-learned features as feedbacks. The incurred clusters are treated as meta-sub-target domains. Hence, our AMEAN auto-designs its multi-target adversarial adaptation loss functions and to this end, dynamically train itself to obtain domain-invariant features from a source to a mixed target and among the multiple meta-sub-target domains derived from the mixed target. This bi-level optimization endows more diverse and flexible adaptation within the mixed target and effectively mitigates its latent category mismatching. 

Our contributions mainly present in three aspects:
\begin{enumerate}[1.]
	\vspace{-4pt}\item{On account of practical cross-domain applications, we consider a new transfer scenario termed Blending-target Domain Adaptation (BTDA), which is common in reality and more difficult to settle.}\vspace{-2pt}
	\item{We propose AMEAN, a adversarial transfer framework incorporating meta-learner dynamically inducing meta-sub-targets to auto-design adversarial adaptation losses, which effectively achieve transfers in BTDA. }\vspace{-2pt}
	\item{Our experiments are conducted on three widely-applied DA benchmarks. Our results show that BTDA setup definitely brings more transfer risks towards existing DA algorithms, while AMEAN significantly outperforms the state-of-the-art and present more robust in BTDA setup. }
\end{enumerate}

\section{Related Work}
Before introducing BTDA problem setup, we would like to briefly revisit (unsupervised) Domain Adaptation (DA) under the modern visual learning background.  

\textbf{Single-source-single-target DA.}
DAs \hspace{-0.1em}are derived from \cite{saenko2010adapting,gopalan2011domain}, where shallow models are deployed to achieve data transfer across visual domains. The development of deep learning enlightens the tunnel to learn nonlinear transferable feature mappings in DAs. Up-to-date deep DA methods have been branched into two mainstreams: explicit and implicit statistical measure matching. The former employs MMD \cite{long2015learning,long2016unsupervised}, CMD \cite{zellinger2017central}, JMMD \cite{long2016deep,Long2017Deep}, \emph{etc}, as the domain regularizer to chase for consistent model performances both on source and target datasets. The latter designates domain discriminators to perform adversarial learning, where the feature extractor is trained to optimize the transferable feature spaces. It includes amounts of avenues to present diverse adversarial manners, \emph{e.g.}, CoGAN \cite{liu2016coupled}, DDC \cite{hoffman2017simultaneous}, RevGred \cite{ganin2014unsupervised,Ganin2017Domain}, ADDA \cite{tzeng2017adversarial}, VADA \cite{shu2018dirt}, GTA \cite{sankaranarayanan2017generate}, \emph{etc}. Beside of the two branches, there are some approaches in virtue of other ideologies, \emph{e.g.}, reconstruction \cite{ghifary2016deep}, semi-supervised learning \cite{saito2017asymmetric} to optimize a domain-aligned feature space.
It is worth noting that, these methods agree with the ``single-source-single-target'' precondition when they learns transferable features for DA . 

\textbf{Multi-source DA (MSDA).} MSDA aims at boosting the target-adaptive accuracy of a model by introducing multiple source domains in a transfer process. \hspace{-0.5em}It is a historical topic \cite{yang2007cross} and refers to a part of DA theories \cite{mansour2009domain} \cite{ben2010theory}. Some recent work place the problem under the deep visual learning background. For instance, \cite{xu2018deep} invented a adversarial reweighting strategy to infer a source-ensemble target classifier; \cite{zhao2018multiple} developed an old-fashion theory to suit deep MSDA and provided a target error upper bound; \cite{mancini2018boosting} stacked multi-DA layers in a network to obtain robust multi-source domain alignment.  

\textbf{Multi-target DA (MTDA).} Similar to MSDA, the goal of MTDA is to enhance data transfer efficacy by bridging the cross-target semantic. \cite{anonymous2019unsupervised} used a semantic disentangler to facilitate an adversarial MTDA approach; \cite{Yu2018Multi} considered the visible semantic gap between multiple targets and propose a dictionary learning algorithm to suit this problem. MTDA is still a fresh area and awaits more explorations.    

 
\section{Blending-target DA (BTDA): Problem Setup}
\textbf{Preliminaries.} Let's consider an $m$-class visual recognition problem. Suppose a source dataset includes $n_s$ labeled examples $\mathcal{S}=\{(\boldsymbol{x}^{(s)}_i,\boldsymbol{y}^{(s)}_i)\}^{n_s}_{i=1}$, where $\boldsymbol{x}^{(s)}_i\in\mathbb{R}_{+}^{d}$ denotes the $i^{th}$ source image lying on a $d$-dimensional data space and $\boldsymbol{y}^{(s)}_i$ is an $m$-dimensional one-hot vector corresponding to its label. Besides, a target set includes $n_t$ unlabeled examples $\mathcal{T}=\{\boldsymbol{x}^{(t)}_i\}^{n_t}_{i=1}$ where $\boldsymbol{x}^{(t)}_i\in\mathbb{R}_{+}^{d}$ denotes the $i^{th}$ target image. $\mathcal{S}$ and $\mathcal{T}$ underly distributions $P_{\mathcal{S}}(\boldsymbol{x},\boldsymbol{y})$ and $P_{\mathcal{T}}(\boldsymbol{x})$, in which $P_{\mathcal{T}}(\boldsymbol{x})=\int P_{\mathcal{T}}(\boldsymbol{x},\boldsymbol{y})d\boldsymbol{y}$ indicates target labels unobservable during training. DA seeks for learning a classifier along with a domain-invariant feature extractor across $\mathcal{S}$ and $\mathcal{T}$, which is capable to predict the correct labels of given images sampled from $P_{\mathcal{T}}(\boldsymbol{x},\boldsymbol{y})$. As of now DA assumes all unlabeled images derived from a single target distribution $P_{\mathcal{T}}$.

Here we turn to consider the multi-target DA (MTDA) setup: every unlabeled target instance $\boldsymbol{x}^{(t)}$ underly $k$ distributions $\{P_{\mathcal{T}_j}(\boldsymbol{x}^{(t)})\}^k_{j=1}$, as they are drawn from the mixture $P_{\mathcal{T}}(\boldsymbol{x}^{(t)})=\sum^{k}_{j=1}\pi_jP_{\mathcal{T}_j}(\boldsymbol{x}^{(t)})$ where $\forall j\in[k], \pi_j\in[0,1] \& \sum^k_j\pi_j=1$. However,
	\emph{The learning goal of MTDA is to 
			simultaneously adapt $k$ targets $\{\mathcal{T}_j\}^k_{j=1}$ instead of their mixed target $\mathcal{T}$.}
Since the multi-target proportions $\{\pi_j\}^k_{j=1}$ are known in MTDA, target set $\mathcal{T}_j$ is explicitly provided by drawing from the posterior $\frac{\pi_{j}P_{\mathcal{T}_j}(\boldsymbol{x}^{(t)})}{\sum^{k}_{j'=1}\pi_jP_{\mathcal{T}_{j'}}(\boldsymbol{x}^{(t)})}$. Hence existing DA algorithms can address the problem by training $k$ target-specific DA models respectively, and using the $j^{th}$-target model to classify the examples\hspace{-0.1em} from\hspace{-0.1em} the $j^{th}$ \hspace{-0.1em}target. 
\begin{figure*}[t]
	\centering
	\includegraphics[height=2.6in]{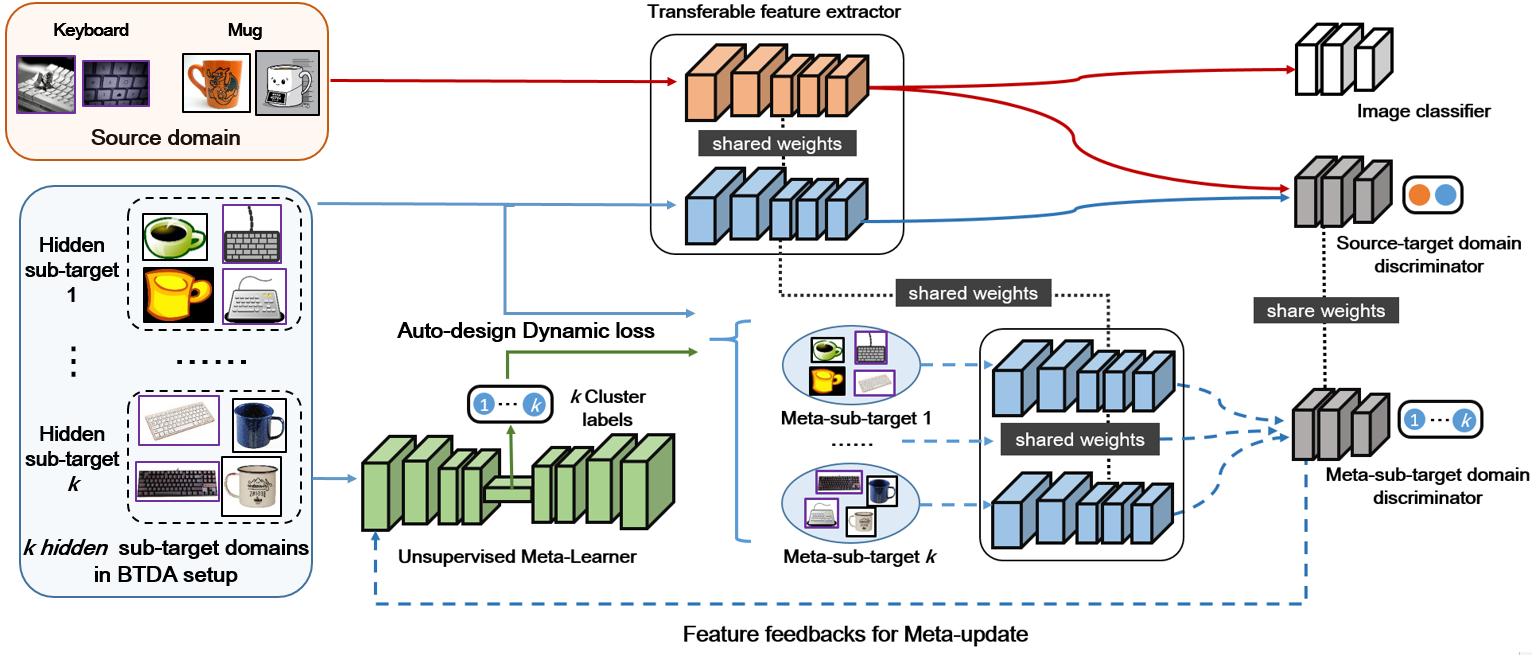}
	\vspace{0pt}\caption{The learning pipeline of our Adversarial MEta-Adaptation Network (AMEAN). AMEAN receives source samples with ground truth and unlabeled samples drawn from an unknown target mixture to synchronously execute two transfer learning processes. In the first process, we propose an adversarial learning bridging our source $\mathcal{S}$ and the mixed target $\mathcal{T}$, so as to learn a feature extractor $F$ that maps them into a common feature space along with a classifier training. In the second stream, AMEAN uses an unsupervised meta-learner $U$ to accept target data and their representations (student feedbacks) and then, learns to separate the mixed target domain into $k$ clusters as meta-sub-targets $\{\mathcal{\hat{T}}_j\}^k_{j=1}$. These meta-sub-targets are iteratively updated to auto-design the multi-target adaptation objectives (Eq.\ref{ddc1} \ref{ddc2}), which operates to progressively eliminate the category mismatching behide the mixed target. Best viewed in color.}\vspace{-14pt} \label{hat}
\end{figure*}

\textbf{From MTDA to BTDA.} Like MTDA, BTDA is also established on a target mixture distribution and expected to adapt $k$ targets $\{\mathcal{T}_j\}^k_{j=1}$. However, multi-target proportions $\{\pi_j\}^k_{j=1}$ in BTDA are unobservable. In other words, \emph{\textbf{BTDA learners are solely provided with a mixed target set $\mathcal{T}$ drawn from a $k$-mixture density and required to classify a mixed target test set drawn from the same mixture}}. If we directly leverage existing DA techniques to transfer category information from $\mathcal{S}$ to $\mathcal{T}$, the learning objective will guide domain-invariant features to adapt a mixed target set $\mathcal{T}$ instead of $k$ target $\{\mathcal{T}_j\}^k_{j=1}$.  Since sub-targets from $k$ distributions could be quite distinct in visual realism, adapting to a mixed target probably would result in drastic category mismatching and arouses serious negative transfer effects. 


\section{Adversarial MEta-Adaptation Network}
To reap positive transfer performance in BTDA setup, we propose Adversarial MEta-Adaptation Network (AMEAN). AMEAN is coupled of two adversarial learning processes, which are parallely executed to obtain domain-invariant features by transferring data from source $\mathcal{S}$ to mixed target $\mathcal{T}$, and among $k$ ``meta-sub-targets'' $\{\mathcal{\hat{T}}_j\}^k_{j=1}$ within the mixed target $\mathcal{T}$. The pipeline of AMEAN is concisely illustrated in Fig.\ref{hat} and we elaborate the methodology as follows.

\subsection{Adaptation from a source to a mixed target}
Suppose that $C$ is a $m$-slot softmax classifier proposed on a transferable feature space and $F$ denotes the feature extractor we long to optimize in BTDA. We employ a source-target domain discriminator $D_{st}$ to deploy an adversarial learning scheme, where $F$ is trained to match a source set $\mathcal{S}$ and a mixed target set $\mathcal{T}$ (the unification of $\{\mathcal{T}_j\}^k_{j=1}$) at the feature level, while $D_{st}$ is demanded to separate the source and the target features. 
	\begin{equation}\begin{aligned}
	&\underset{F,C}{\min} \ \underset{D_{st}}{\max} \ V_{\rm st}(F,D_{st},C)= \\&  \lambda\Big(\underbrace{\mathbb{E}_{\boldsymbol{x}\sim\mathcal{S}}\log \big[D_{st}\big(F(\boldsymbol{x})\big)\big]+\mathbb{E}_{\boldsymbol{x}\sim\mathcal{T}}\log \big[1-D_{st}\big(F(\boldsymbol{x})\big)\big]}_{\rm Adversarial \ DA \ loss}\Big)\\ &\ \ \ \ \ \ \ \ \ \ \ \ \ \ \ \ \ \ \ \ \ \ \ \ \ \ \ \ \ \ \ \ \ \ \ \ \ \ \ \ \ \ \  \underbrace{-\mathbb{E}_{(\boldsymbol{x},\boldsymbol{y})\sim\mathcal{S}} \ \boldsymbol{y}^T\log \ C\big(F(\boldsymbol{x})\big)}_{\rm Classification \ loss}\label{eq1.1}
	\end{aligned}
	\end{equation}. This minimax objective can be optimized in two ways: 
\begin{enumerate}
	\vspace{-6pt}\item $C$, $F$, $D_{st}$ are jointly trained by inducing a reversed gradient layer \cite{ganin2014unsupervised} \cite{Ganin2017Domain} across $D_{st}$ and $F$, which inversely back-propagates the gradients from a source-target domain discriminator $D_{st}$.\vspace{-6pt}
	\item $C$, $F$ and $D_{st}$ are updated by an alternative optimization 
		 like GAN \cite{Goodfellow2014Generative} \cite{shu2018dirt}.\vspace{-2pt}
\end{enumerate}Our experiments show them both well-suited in AMEAN.
\subsection{Meta-adaptation among meta-sub-targets}
The pure source-to-target transfer is blind to the domain gaps and category misalignment among $\{\mathcal{T}_j\}^k_{j=1}$, which the negative transfer mainly ascribes to. Here we interpret the key module of AMEAN to address this problem: an unsupervised meta-learner trained to obtain $k$ embedded clusters \cite{xie2016unsupervised} as our ``meta-sub-targets'', which play the roles of automatically and dynamically outputting multi-target adversarial adaptation loss functions  (Eq.\ref{ddc1}, \ref{ddc2}) in pursuit of reducing the category mismatching inside the mixed target $\mathcal{T}$.   

\textbf{Meta-learning for dynamic loss design.}. Meta-learning (``learning to learn'') \cite{andrychowicz2016learning} become a thriving topic in machine learning community. It refers to all learnable technique to improve model generalization, which includes fast adaptation to rare categories \cite{Wang2018Low} and unseen tasks \cite{Finn2017Model}, as well as auto-tuning the hyper-parameters accompanied with training, \emph{e.g.}, learning rate \cite{andrychowicz2016learning}, architecture \cite{zoph2016neural}, loss function \cite{Xu2018AutoLoss}, \emph{etc}. Inspired by them, our meta-learner $U$ is an unsupervised net learning to find deep clustering embeddings that incurs $k$ clusters based on the ongoing feature extractor feedbacks. It induces $k$ meta-sub-targets $\{\mathcal{\hat{T}}_j(U)\}^k_{j=1}$ to take place of $\{\mathcal{T}_j\}^k_{j=1}$. Via this auxiliary, meta-learner has access to auto-design the adversarial multi-target DA loss with respect to the $k$ meta-sub-targets and, dynamically alter the losses according to the change of these meta-sub-targets as our meta-learner updates the $k$ clusters. 
	
Two major concerns are probably raised to our methodology: 1). Are $\{\mathcal{\hat{T}}_j(U)\}^k_{j=1}$ and $\{\mathcal{T}_j\}^k_{j=1}$ similar ? 2). Does the clustering finally lead to target samples of the same category staying together? Towards the first concern, our answer is not and unnecessary. The technical difficulty in BTDA arises from the category misalignment instead of the hidden sub-target domains. As long as a DA algorithm performs appropriate category alignment in a mixed target, it is unnecessary to explicitly discover these hidden sub-target domains. AMEAN receives ongoing learned features that implicitly conceive label information from Eq \ref{eq1.1} , to adaptively organizes its meta-adaptation objective. In our ablation, This manner shows powerful to overcome the category misalignment in $\mathcal{T}$. 
The second concern raises a dilemma to our meta-transfer: as learned features become more classifiable, target features of the same class will be closer and closer, then clustering will more probably select them as a newly-updated meta-target domain. It causes category shift if we apply them to learn AMEAN. To remove this hidden threat, our meta-learner simultaneously receives $\boldsymbol{x}^{(t)}$ and $F(\boldsymbol{x}^{(t)})$ to learn deep clustering embeddings, where $\boldsymbol{x}^{(t)}$ denotes the primitive state of a sample without DA and $F(\boldsymbol{x}^{(t)})$ implies the adaptation feedback from the ongoing learned feature. Since $\boldsymbol{x}^{(t)}$ is feature-agnostic, the induced clusters inherit the merit of meta-learning and concurrently gets rid of this classifiable feature dilemma. 


\vspace{-6pt}\textbf{Discovering $k$ meta-sub-targets via deep $k$-clustering.} Our unsupervised meta-learner $U$ is derived from the base model in DEC \cite{xie2016unsupervised}, a denoising auto-encoder (DAE) composed of encoder $U_1$ and decoder $U_2$. Distinguished from the DAE in DEC, $U$ takes a target couple $
\big(\boldsymbol{x}^{t},F(\boldsymbol{x}^{(t)})\big)
$ as inputs and learns to satisfy the self-reconstruction as well as obtain deep clustering embedding $U_1\big(\boldsymbol{x}^{(t)},F(\boldsymbol{x}^{(t)})\big)$. More specifically, suppose $\{\mu_j\}^k_{j=1}$ is the $k$ cluster centroids, and we define a soft cluster assignment $\{q_{i,j}\}^{k}_{j=1}$ to $\boldsymbol{x}^{(t)}_i\in\mathcal{T}$.
\begin{equation}
q_{i,j} = \frac{(1+\frac{||U_1(\boldsymbol{x}^{(t)}_i,F(\boldsymbol{x}^{(t)}_i))-\mu_j||^2}{\alpha})^{-\frac{\alpha+1}{2}}}{\sum_{j'}^{k}(1+\frac{||U_1(\boldsymbol{x}^{(t)}_i,F(\boldsymbol{x}^{(t)}_i))-\mu_{j'}||^2}{\alpha})^{-\frac{\alpha+1}{2}}}\label{q}
\end{equation}where $\alpha$ indicates the degree of freedom in a Student's \emph{t}-distribution. Eq.\ref{q} is aimed to learn $U$ by iteratively inferring deep cluster centroids $\{\mu_j\}^k_{j=1}$. This EM-like learning operates with the help of auxiliary distributions $\{p_{i,j}, \forall i\in[n_t]\}$, which are computed by raising $q_i$ to the second power and normalize it by the frequency in per cluster, \emph{i.e.},
\begin{small}
	\begin{equation}
	p_{i,j}=\frac{q^2_{i,j}/f_j}{\sum_{j'}^{k}q^2_{i,j'}/f_{j'}}, \ \ s.t. \ f_j=\sum_{i}^{n_t}q_{i,j}\label{p}
	\end{equation}
\end{small}where $f_j$ denotes the $j^{th}$ cluster frequency. Compared with $q_i$, $p_i$ endows more emphasis on data points with high confidence and thus, is more appropriate to supervise the soft cluster inference. We employ KL divergence to restrict $q_{i}$ and $p_{i}$ for the meta-learner clustering network learning: 
\begin{equation}
\begin{aligned}
\underset{U_1, U_2, \{\mu_j\}^k_{j=1}}{\min} \ \mathbb{E}_{\boldsymbol{x}_i\sim \mathcal{T}} \ L_{\rm rec}(\boldsymbol{x}_i;F)-\sum_{j=1}^{k} p_{i,j}\log\frac{p_{i,j}}{q_{i,j}}\label{U}
\end{aligned}
\end{equation}where $L_{\rm rec}(\boldsymbol{x};F)$ denotes a $l_2$ self-reconstruction \emph{w.r.t.} a target feedback pair $\big(\boldsymbol{x},F(\boldsymbol{x})\big)$ and the second term denotes a KL divergence term for clustering. Parameter learning and $k$ cluster centroid ($\{\mu_j\}^k_{j=1}$) update are facilitated by back-propagation with a SGD solver. (See more in Appendix.A) 

After meta-learner $U$ converges, we apply the incurred clustering assignments to separate $\mathcal{T}$ into $k$ meta-sub-target domains, \emph{i.e.}, $\boldsymbol{x}^{(t)}_i$ in a mixed target $\mathcal{T}$ will be classed into meta-sub-target $\mathcal{\hat{T}}_j(U)$ if $q_{i,j}$ is the maximum in $\{q_{i,j'}\}^k_{j'=1}$:  
\begin{equation}
\forall j\in[k], \ \ \mathcal{\hat{T}}_j(U)=\{\boldsymbol{x}_i\in\mathcal{T} \& j=\underset{j'}{\arg\max} \ q^\ast_{i,j'} \}\label{split}
\end{equation}
\textbf{Meta-sub-target adaptation.} Given $k$ meta-sub-target domains $\{\mathcal{\hat{T}}_j(U)\}^k_{j=1}$, AMEAN auto-designs the $k$-sub-target DA losses to re-align the features in $\mathcal{T}$. More detailedly, AMEAN designates a $k$-slot softmax classifier $D_{mt}$ sharing parameters with $D_{st}$, as meta-sub-target domain discriminator. In order to obtain $k$ meta-sub-target adaptations, features extracted by $F$ seek to ``maximally confuse'' the discriminative decision of $D_{mt}$: 
	\begin{small}
		\begin{equation}
		\begin{aligned}
		\underset{F}{\min} \ \underset{D_{\rm mt}}{\max} \ V_{\rm mt}(F,D_{\rm mt}) = \sum_{j=1}^{k}\mathbb{E}_{\boldsymbol{x}\sim\hat{\mathcal{T}}_j(U)}\boldsymbol{1}_j^T\log \big[D_{mt}\big(F(\boldsymbol{x})\big)\big]\label{ddc1}
		\end{aligned}
		\end{equation}
	\end{small}where $\boldsymbol{1}_j$ indicates a $k$-dimensional one-hot vector implying that sample $\boldsymbol{x}$ belongs to $\mathcal{\hat{T}}_j(U)$. In the case of joint parameter learning, due to the mutual architectures of $D_{st}$ and $D_{mt}$, Eq.\ref{ddc1} is implemented by the same reversed gradient layer originally for source-to-target transfer. However, if subnets $F$ and $D_{\rm mt}$ are alternatively trained, Eq.\ref{ddc1} is solely used to update $D_{\rm mt}$ while we prefer to optimize feature extractor $F$ by maximizing the cross-entropy of $D_{mt}\big(F(\boldsymbol{x})\big)$:
	\begin{small}
			\begin{equation}\begin{aligned}
			\underset{F}{\min} \ \widetilde{V}_{\rm mt}(F,D_{mt}) = \sum_{j=1}^{k}\mathbb{E}_{\boldsymbol{x}\sim\hat{\mathcal{T}}_j(U)}D_{mt}\big(F(\boldsymbol{x})\big)^T\log \big[D_{mt}\big(F(\boldsymbol{x})\big)\big]\label{ddc2}
			\end{aligned}
			\end{equation}
	\end{small} It implies that $F$ learns to ``confuse'' multi-target domain discriminator $D_{\rm mt}$, namely, $D_{\rm mt}$ could not identify which meta-sub-target an unlabeled example belongs to.
	
\begin{algorithm}[t]			
	\caption{AMEAN (Stochastic version)} \label{A1}
	\begin{small}
		\hspace*{0.02in}{\bf Input:} Source $\mathcal{S}$; Mixed target $\mathcal{T}$; feature extractor $F$; classifier $C$; domain discriminators $D_{st}$,$D_{mt}$; meta-learner $U=\{U_1,U_2\}$.\\
		\hspace*{0.02in}{\bf Output:} well-trained $F^\ast$, $D_{st}^\ast$,$D_{mt}^\ast$, $C^\ast$.
		\begin{algorithmic}[1]
			\While{not converged}
			\State \textbf{\emph{Meta-sub-target Discovery} (Meta-update):}
			\State Initiate $\{\mu_j\}^k_{j=1}$ and $U$;
			\While{not converged}
			\State Sample a mini-batch $X_t$ from $\mathcal{T}$; Construct $\{q_{i,j}\}^{k}$ by Eq.\ref{q} and $\{p_{i,j}\}^{k}$ by Eq.\ref{p} for each $\boldsymbol{x}_i$ in the mini-batch.
			\State Update $U$ and $\{\mu_j\}_{j=1}^{k}$ in Eq.\ref{U} with SGD solver.
			\EndWhile
			\State Divide $\mathcal{T}$ into $k$ meta-sub-targets $\{\mathcal{\hat{T}}_j(U)\}^k_{j=1}$ by Eq.\ref{split}.
			\State \textbf{\emph{Collaborative Advesarial Meta-Adaptation}:}
			\For{1:$M$}
			\State Sample a mini-batch $X_s$ from $\mathcal{S}$ and $k$ mini-batches $\{X^{(j)}_t\}^k_{j=1}$ from $\{\mathcal{\hat{T}}_j(U)\}^k_{j=1}$ respectively; $X_t=\cup X^{(j)}_t$.
			\If {alternating domain adaptation} \State Update $D_{st}$ and $D_{mt}$ with Eq.\ref{eq.alt1} .
			\State Update $F$ and $C$ with Eq.\ref{eq.alt2} .
			\Else \ \ Update $D_{st}$, $D_{mt}$, $F$ and $C$ with Eq.\ref{eq.joint} .
			\EndIf 
			\EndFor
			\EndWhile \\
			\Return $F^\ast = F$; $C^\ast = C$; $D_{st}^\ast = D_{st}$; $D_{mt}^\ast=D_{mt}$.
		\end{algorithmic}\label{algo}
	\end{small}\vspace{-2pt}
\end{algorithm}

\begin{table*}[h]
	\centering
	\caption{ Classification accuracy (ACC \%)on Digit-five in BTDA setup. \textbf{{\color{blue}BLUE}}, \textbf{{\color{red}RED}} indicate the baseline suffer from \emph{absolute negative transfer} (ANT\%) or \emph{relative negative transfer} (RNT\%), respectively. Best viewed in color. }\label{t1}
	\begin{scriptsize}
		\begin{tabular}{|l|cc|cc|cc|cc|cc|cc|r}\thickhline
			Models &\multicolumn{2}{c|}{mt$\rightarrow$mm,sv,up,sy} &\multicolumn{2}{c|}{mm$\rightarrow$mt,sv,up,sy} &\multicolumn{2}{c|}{sv$\rightarrow$mm,mt,up,sy} &\multicolumn{2}{c|}{sy$\rightarrow$mm,mt,sv,up} &\multicolumn{2}{c|}{up$\rightarrow$mm,mt,sv,sy} &\multicolumn{2}{c|}{Avg}\\
			&ACC$^{\rm ANT}$		&RNT	 
			&ACC$^{\rm ANT}$		&RNT	 
			&ACC$^{\rm ANT}$		&RNT	 
			&ACC$^{\rm ANT}$		&RNT	 
			&ACC$^{\rm ANT}$		&RNT	 
			&ACC$^{\rm ANT}$		&RNT	 
			\\		\hline
			\textbf{Backbone-1:}	&	&	&	&	& &	 &  &	 &  & &  &	\\
			Source only				&26.9 &0	&56.0 &0  	&67.2 &0	 &73.8 &0 &36.9 &0 &52.2 &0 		\\
			ADDA		&43.7 &{\color{red}-8.0}	&{\color{blue}55.9$^{(-0.1)}$}&{\color{red}-3.3}	&{\color{blue}40.4$^{(-26.8)}$}&{\color{red}-21.7}	 &{\color{blue}66.1$^{(-6.7)}$}&{\color{red}-6.5} &{\color{blue}34.8$^{(-0.1)}$}&{\color{red}-13.3} &{\color{blue}48.2$^{(-4.0)}$}&{\color{red}-10.5}		\\
			DAN				&31.3&{\color{red}-7.5}	 &{\color{blue}53.1$^{(-2.9)}$}&{\color{red}-3.1}&{\color{blue}48.7$^{(-18.5)}$}&-{\color{red}9.5}&{\color{blue}63.3$^{(-10.5)}$}&{\color{red}-3.9} &{\color{blue}27.0$^{(-9.9)}$}&{\color{red}-11.0} 	&{\color{blue}44.7$^{(-7.5)}$}&{\color{red}-7.0}		\\
			GTA				&44.6&{\color{red}-9.2}	&{\color{blue}54.5$^{(-1.5)}$}&{\color{red}-2.1}	  	&{\color{blue}60.3$^{(-6.9)}$}&{\color{red}-3.9} &\textbf{74.5 (+0.7)}&{\color{red}-1.1}&41.3&{\color{red}-2.0}&55.0&{\color{red}-3.7}		\\
			RevGrad		&52.4&{\color{red}-8.9}&64.0&{\color{red}-4.1}&{\color{blue}65.3$^{(-1.9)}$}&{\color{red}-4.1}&{\color{blue}66.6$^{(-6.2)}$}&{\color{red}-7.5}	  	&44.3&{\color{red}-6.3} &58.5&{\color{red}-6.2} 		\\
			AMEAN   &\textbf{56.2 (+3.8)}&-	&\textbf{65.2 (+1.2)}&-	&\textbf{67.3 (+0.1)}	&-  	&{\color{blue}71.3$^{(-2.5)}$}	&-	 &\textbf{47.5 (+3.2)}  &- &\textbf{61.5 (+3.0)} &-\\
			\hline
			\textbf{Backbone-2:}	&	&	&	&	& &	 &  &	 &  & &  &	\\
			Source only				&47.4 &0	&58.1 &0  	&73.8 &0 &74.5 &0 &50.6 &0  &60.8 &0 		\\
			VADA				&76.0&{\color{red}-5.3}	  	&72.3&{\color{red}-2.3} &75.6&{\color{red}-2.5}&81.3&{\color{red}-3.8}&56.4&{\color{red}-8.7} &72.3&{\color{red}-4.5} 		\\
			DIRT-T				&73.5&{\color{red}-7.1} &76.1&{\color{red}-1.5}&75.9&{\color{red}-5.5}&78.5&{\color{red}-3.1}&47.0&{\color{red}-7.5}&70.2&{\color{red}-5.0} 		\\
			AMEAN  &\textbf{85.1 (+9.1)}&-	&\textbf{77.6 (+1.5)} &-	&\textbf{77.4 (+1.5)}	&-  	&\textbf{84.1 (+2.8)}	&-	 &\textbf{75.5 (+19.1)}  &- &\textbf{80.0 (+7.7)} &-\\
			\hline
		\end{tabular}
	\end{scriptsize}\vspace{-12pt}
\end{table*}
\subsection{Collaborative Advesarial Meta-Adaptation}
In order to learn domain-invariant features as well as resist the negative transfer from a mixed target, the previous transfer processes should be combined to battle the domain shifts. In particular, we retrain our meta-learner to update meta-DA losses $V_{\rm mt}(F, D_{\rm mt})$ and $\widetilde{V}_{\rm mt}(F, D_{\rm mt})$ per $M$ iteration during feature learning. After that, if we employ a reverse gradient layer as the adversarial DA implementation (Eq.\ref{eq1.1}), the collaborative learning objective is formulated as  
\begin{equation}
\begin{aligned}
\underset{D_{\rm st}, D_{\rm mt}}{\max} \underset{F, C}{\min} \ \ &V_{\rm joint}(D_{\rm st},D_{\rm mt},F, C) \\&= V_{\rm st}(F, D_{\rm st}, C) +\gamma V_{\rm mt}(F, D_{\rm mt}) \label{eq.joint}
\end{aligned}
\end{equation}where $\gamma$ indicates the balance factor between two transfer processes. Eq.\ref{eq.joint} suits joint learning \emph{w.r.t.} $F$, $D_{\rm st}$, $D_{\rm mt}$, $C$, while in an alternating adversarial manner, it would be more appropriate to iteratively update $\{F,C\}$ and $\{D_{\rm st}, D_{\rm mt}\}$ by switching the optimization objectives between
\begin{equation}
\begin{aligned}
\underset{D_{\rm st}, D_{\rm mt}}{\max} \ V_{\rm alter}(D_{\rm st},D_{\rm mt}) = V_{\rm st}(F, D_{\rm st}, C)+V_{\rm mt}(F, D_{\rm mt})\label{eq.alt1}
\end{aligned}
\end{equation}
\begin{equation}
\begin{aligned}
\underset{F, C}{\min} \ V_{\rm alter}(F, C) = V_{\rm st}(F, D_{\rm st}, C)+\gamma \widetilde{V}_{\rm mt}(F, D_{\rm mt})\label{eq.alt2}
\end{aligned}
\end{equation}In a summary, the stochastic learning pipeline of AMEAN is described by Algorithm.\ref{algo} .
	
\vspace{-6pt}\section{Experiments}\vspace{-4pt}
In this section, we elaborate comprehensive experiments in the BTDA setup and compare AMEAN with state-of-the-art DA baselines.   
\subsection{Setup}\vspace{-2pt}
\textbf{Benchmarks.} \hspace{-0.4em}
\emph{\textbf{Digit-five}} \cite{xu2018deep} is composed of five domain \hspace{-0.2em}sets drawn from \textbf{mt} (\emph{MNIST}) \cite{lecun1998gradient}, \textbf{mm} (\emph{MNIST-M}) \cite{Ganin2017Domain}, \textbf{sv}(\emph{SVHN}) \cite{Netzer2011Reading}, \textbf{up} (\emph{USPS}) and \textbf{sy} (\emph{Synthetic Digits}) \cite{Ganin2017Domain}, respectively. There are 25000 for training and 9000 for testing in \textbf{mt}, \textbf{mm}, \textbf{sv}, \textbf{sy}, while the entire \emph{USPS} is chosen as a domain set \textbf{up}.
\emph{\textbf{Office-31}} \cite{saenko2010adapting} is a famous visual recognition benchmark comprising 31 categories and totally 4652 images in three separated visual domains \textbf{A} (\emph{Amazon}), \textbf{D} (\emph{DSLR}), \textbf{W} (\emph{Webcam}), which indicate images taken by web camera and digital camera in distinct environments.
\emph{\textbf{Office-Home}} \cite{venkateswara2017deep} consists of four visual domain sets, i.e., Artistic (\textbf{Ar}), Clip Art (\textbf{Cl}), Product (\textbf{Pr}) and Real-world (\textbf{Rw}) with 65 categories and around $15,500$ images in total. 

\textbf{Baselines.} Since MSDA, MTDA approaches require domain remarks, \hspace{-0.2em}they obviously do not suit the BTDA setup. Therefore we compare our AMEAN with existing (\emph{single-source-single-target}) DA baselines and evaluate their classification accuracies by transferring class information from source $\mathcal{S}$ to mixed target $\mathcal{T}$. State-of-the-art DA baselines include: Deep Adaptation Network (\textbf{DAN}) \cite{long2015learning}, Residual Transfer Network (\textbf{RTN}) \cite{long2016unsupervised}, Joint Adaptation Network (\textbf{JAN}) \cite{Long2017Deep}, Generate To Adapt (\textbf{GTA}) \cite{sankaranarayanan2017generate}, Adversarial Discriminative Domain Adaptation (\textbf{ADDA}) \cite{tzeng2017adversarial}, Reverse Gradient (\textbf{RevGred}) \cite{ganin2014unsupervised} \cite{Ganin2017Domain}, Virtual Adversarial Domain Adaptation (\textbf{VADA}) \cite{shu2018dirt} and its variant \textbf{DIRT-T} \cite{shu2018dirt}. DAN, RTN and JAN proposed MMD-based regularizer to pursue cross-domain distribution matching in a feature space; ADDA, RevGred, GTA and VADA are domain adversarial training paradigms encouraging domain-invariant feature learning by ``cheating'' their domain discriminators. DIRT-T is built upon VADA by introducing a network to guide the dense target feature regions away from the decision boundary. Beyond these approaches, we also report the \textbf{Source-only} results based on $F$ and $C$ that are merely trained on source labeled data to classify target examples. 
\begin{table*}[h]
	\centering
	\caption{Classification accuracy (ACC \%) on Office-31 in BTDA setup. \textbf{{\color{blue}BLUE}}, \textbf{{\color{red}RED}} indicate the baseline suffer from \emph{absolute negative transfer} (ANT\%) or \emph{relative negative transfer} (RNT\%), respectively. Best viewed in color.}\label{t2}
	\begin{footnotesize}
		\begin{tabular}{|lc|cc|cc|cc|cc|r}\thickhline
			\multirow{2}{*}{Backbones}&\multirow{2}{*}{Models} &\multicolumn{2}{c|}{A$\rightarrow$D,W} &\multicolumn{2}{c|}{D$\rightarrow$A,W} &\multicolumn{2}{c|}{W$\rightarrow$A,D} &\multicolumn{2}{c|}{Avg}\\
			&&ACC$^{\rm ANT}$		&RNT	&ACC$^{\rm ANT}$		&RNT	&ACC$^{\rm ANT}$		&RNT	&ACC$^{\rm ANT}$		&RNT			\\\hline
			\multirow{6}{*}{AlexNet}&Source only	 &62.4 &0		&60.8 &0		&57.2 &0	&60.1 &0	\\	
			&DAN				&68.2&{\color{red}-0.2}	&{\color{blue}58.7$^{(-2.1)}$}&{\color{red}-5.0}		&{\color{blue}55.6$^{(-1.6)}$}&{\color{red}-4.9}&60.8&{\color{red}-3.4}		\\
			&RTN				&71.6&{\color{red}-1.1}		&{\color{blue}56.3$^{(-4.5)}$}&{\color{red}-4.6}		 &{\color{blue}52.2$^{(-5.0)}$}&{\color{red}-6.1} &{\color{blue}59.9$^{(-0.2)}$}&{\color{red}-4.1}		\\
			&JAN				&73.7&{\color{red}-0.3}		&62.1&{\color{red}-4.9}		 &58.4&{\color{red}-3.6} &64.7&{\color{red}-3.0}		\\
			&RevGrad				&74.1&+0.9	 &{\color{blue}58.6$^{(-2.2)}$}&{\color{red}-4.3}	  	&{\color{blue}55.0$^{(-2.2)}$}&{\color{red}-3.4}&62.6&{\color{red}-2.2}		\\
			&AMEAN (ours)  	&\textbf{74.5 (+0.4)}	&-  	&\textbf{62.8 (+0.7)}	&-	 &\textbf{59.7 (+1.3)}  &- &\textbf{65.7 (+1.0)} &-\\
			\hline
			\multirow{6}{*}{ResNet-50}&Source only	 &68.6 &0		&70.0 &0		&66.5 &0	&68.4 &0	\\	
			&DAN				&78.0&{\color{red}-2.1}	&{\color{blue}64.4$^{(-5.6)}$}&{\color{red}-6.8}		&66.7&{\color{red}-1.8}&69.7&{\color{red}-3.6}		\\
			&RTN				&84.3&+2.3		&{\color{blue}67.5$^{(-2.5)}$}&{\color{red}-5.5}		 &{\color{blue}64.8$^{(-0.2)}$}&{\color{red}-5.5} &72.2&{\color{red}-2.9}		\\
			&JAN				&84.2&{\color{red}-1.2}		&74.4&{\color{red}-0.8}		 &72.0&{\color{red}-2.8} &76.9&{\color{red}-1.6}		\\
			&RevGrad				&78.2&{\color{red}-3.3}	 &72.2&{\color{red}-2.7}	  	&69.8&{\color{red}-2.8}&73.4&{\color{red}-2.9}		\\
			&AMEAN (ours)  	&\textbf{90.1 (+5.8)}	&-  	&\textbf{77.0 (+2.6)}	&-	 &\textbf{73.4 (+1.4)}  &- &\textbf{80.2 (+3.4)} &-\\
			\hline
		\end{tabular}
	\end{footnotesize}
	\center\vspace{-4pt}
	\caption{ Classification accuracy (ACC \%) on Office-Home in BTDA setup. \textbf{{\color{blue}BLUE}}, \textbf{{\color{red}RED}} indicate the baseline suffer from \emph{absolute negative transfer} (ANT\%) or \emph{relative negative transfer} (RNT\%), respectively. Best viewed in color.}\label{t3}\vspace{-0pt}
	\begin{footnotesize}
		\begin{tabular}{|lc|cc|cc|cc|cc|cc|r}\thickhline
			\multirow{2}{*}{Backbones}&\multirow{2}{*}{Models} &\multicolumn{2}{c|}{Ar$\rightarrow$Cl,Pr,Rw} &\multicolumn{2}{c|}{Cl$\rightarrow$Ar,Pr,Rw} &\multicolumn{2}{c|}{Pr$\rightarrow$Ar,Cl,Rw} &\multicolumn{2}{c|}{Rw$\rightarrow$Ar,Cl,Pr} &\multicolumn{2}{c|}{Avg}\\
			&&ACC$^{\rm ANT}$		&RNT	&ACC$^{\rm ANT}$		&RNT	&ACC$^{\rm ANT}$		&RNT	&ACC$^{\rm ANT}$		&RNT	&ACC$^{\rm ANT}$		&RNT		\\\hline
			\multirow{6}{*}{AlexNet}&Source only				&33.4 &0 &37.6 &0 &32.4 &0&39.3 &0&35.7 &0 		\\
			&DAN				&39.7&{\color{red}-3.7}	 &43.2&{\color{red}-3.0}	&39.4&{\color{red}-3.4}&47.8&{\color{red}-2.2}&42.5&{\color{red}-3.1}	\\
			&RTN				&42.8&{\color{red}-2.0} &45.2&{\color{red}-2.5} &40.6&{\color{red}-2.4}	  			 &49.6&{\color{red}-2.7}  &44.6&{\color{red}-2.3}		\\
			&JAN				&43.5&{\color{red}-2.9} &46.5&{\color{red}-3.6} &40.9&{\color{red}-6.6}	  			 &49.1&{\color{red}-2.1}  &45.0&{\color{red}-4.6}		\\
			&RevGrad				&42.1&{\color{red}-3.3}& 45.1&{\color{red}-4.4}		&41.1&{\color{red}-4.5}	  	&48.4&{\color{red}-5.6} &44.2&{\color{red}-4.4}		\\
			&AMEAN (ours)  	&\textbf{44.6 (+1.1)}	&-  &\textbf{47.6 (+1.1)}	&-	 &\textbf{42.8 (+1.7)}  &- &\textbf{50.2 (+1.1)} & &\textbf{46.3 (+1.3
				)} &-\\
			\hline
			\multirow{6}{*}{ResNet-50}&Source only				&47.6 &0 &42.6 &0 &44.2 &0&51.3 &0&46.4 &0 		\\
			&DAN				&55.6&{\color{red}-0.8}	 &56.6&+0.9	&48.5&{\color{red}-5.1}&56.7&{\color{red}-6.3}&54.4&{\color{red}-2.6}	\\
			&RTN				&53.9&{\color{red}-1.8} &56.7&{\color{red}-1.3} &47.3&{\color{red}-3.8}	  			 &51.6&{\color{red}-2.8}  &52.4&{\color{red}-2.6}		\\
			&JAN				&58.3&{\color{red}-0.4} &60.5&+2.3 &52.2&{\color{red}-2.2}	  			 &57.5&{\color{red}-7.0}  &57.1&{\color{red}-1.9}		\\
			&RevGrad				&58.4&{\color{red}-3.1}& 58.1&{\color{red}-2.2}		&52.9&{\color{red}-4.5}	  	&62.1&{\color{red}-3.0} &57.9&{\color{red}-3.2}		\\
			&AMEAN (ours)  	&\textbf{64.3 (+5.9)}	&-  &\textbf{65.5 (+5.0)}	&-	 &\textbf{59.5 (+6.1)}  &- &\textbf{66.7 (+4.6)} & &\textbf{64.0 (+6.1)} &-\\
			\hline
		\end{tabular}
	\end{footnotesize}\vspace{-12pt}
\end{table*}

\textbf{Implementation setting.} In digit recognition, we evaluate AMEAN on two different backbones. The first is derived from a LeNet architecture with $F$, $D_{\rm st}$, $D_{\rm mt}$, $C$ jointly trained through a reversed gradient layer (Eq.\ref{eq.joint}); the second employs a GAN-based alternating learning scheme \cite{Goodfellow2014Generative} that switches the optimization between Eq.\ref{eq.alt1} , \ref{eq.alt2} . For a fair comparison, all baselines in Digit-five experiment are based on these backbones. 
In Office-31 and Office-Home, we evaluate all baselines with AlexNet\cite{krizhevsky2012imagenet} and ResNet-50 \cite{He2015Deep}, where our AMEANs are trained by Eq.\ref{eq.joint}. Our meta-learner employ the same architecture in all experiments, \emph{i.e.}, a four-layered fully-connected DAE. 
More implementation details are deferred in our Appendix.A. 

\subsection{Evaluation Criteria.}\vspace{-6pt} 
Our experimental evaluation is aimed to answer two fundamental questions in this paper:
\begin{enumerate}
	\vspace{-4pt}\item \textbf{\emph{Does BTDA bring more transfer learning risks to existing DA algorithms}} ?\vspace{-4pt}
	\item \textbf{\emph{Is our AMEAN able to reduce these transfer risks}} ?
\end{enumerate}\vspace{-6pt}
As a primal metric, classification accuracies on a mixed target ($Acc_{\rm \textbf{BTDA}}$) are provided to evaluate DA baselines in the BTDA setup, where their adaptations from $\mathcal{S}$ to a mixed $\mathcal{T}$ are performed to cultivate a classifier that predicts labels on a mixed target test set. To answer the questions above, we also consider two additional metrics, \emph{i.e.}, \emph{absolute negative transfer} (ANT) and \emph{relative negative transfer} (RNT):
\begin{itemize}
	\item \textbf{Absolute negative transfer (ANT).} Given a DA baseline, if its performance is inferior to its Source-only, it implies that this DA algorithm not only fails to benefit but also damages the classifier, \emph{i.e.}, suffers from ANT.
	\vspace{-16pt}
	\item \textbf{Relative negative transfer (RNT).} RNT aims to measure how much performance drops if a DA baseline alters from MTDA to BTDA setups. In MTDA, each DA baseline performs adaptation from source $\mathcal{S}$ to each explicit target $\mathcal{T}_j$ ($\forall j\in[k]$). It results in 
	$k$ target-specific domain-adaptive models with their accuracies $\{Acc_{j}\}^k_{j=1}$ on the target test sets respectively. Towards this end, we compute the MTDA weighted averaged accuracy by 
	\begin{small}
		$
		Acc_{\rm \textbf{MTDA}} = \sum^k_{j=1} \alpha_j\ Acc_{j}\label{wb}
		$
	\end{small}where $\alpha_j$ indicates the ratio of sub-target $\mathcal{T}_j$ in a mixed target $\mathcal{T}$. Hence we have RNT $=Acc_{\rm \textbf{BTDA}}-Acc_{\rm \textbf{MTDA}}$.\vspace{-6pt}   
\end{itemize} More details of the metrics can be found in our Appendix.B.

\vspace{-4pt}\subsection{Results.}\vspace{-8pt}
The evaluations based on classification accuracy (ACC, \emph{i.e.}, $Acc_{\rm BTDA}$), ANT and RNT have been conducted in Tables.\ref{t1} - \ref{t3} (we highlight ANT, RNT in \textbf{{\color{blue}BLUE}}, \textbf{{\color{red}RED}}). 
\begin{figure*}[t]
	\centering
	\vspace{-16pt}\includegraphics[width=6.4in,height=1.4in]{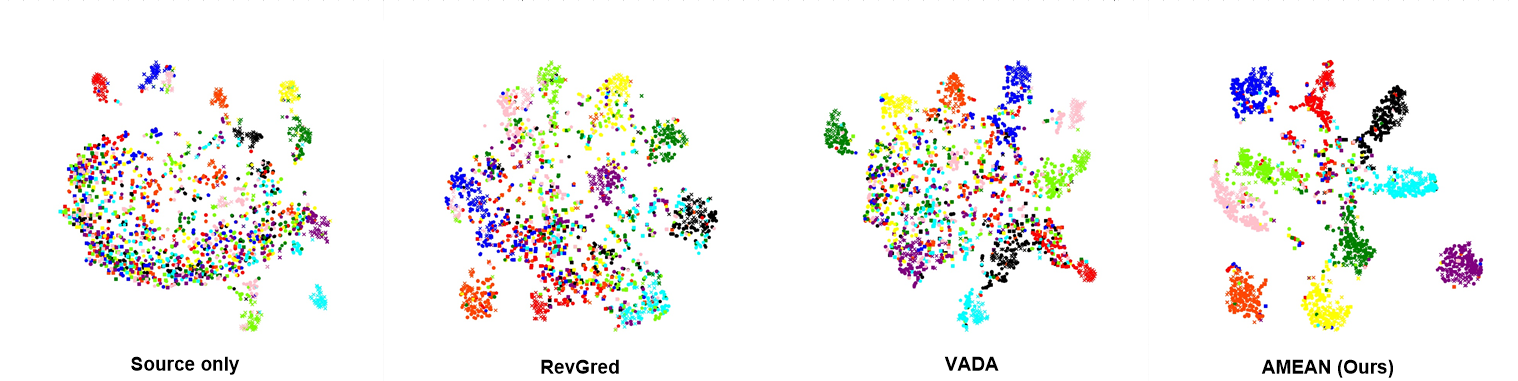}
	\caption{t-SNE visualizations of the features learned by Source-only, RevGred, VADA and AMEAN on Digit-five in BTDA setup. Shapes and colors indicate different domains and categories, respectively. Best viewed in color. } \label{vis}\vspace{-12pt}
\end{figure*}
\begin{figure}[t]
	\centering
	\vspace{-6pt}\includegraphics[height=1.55in]{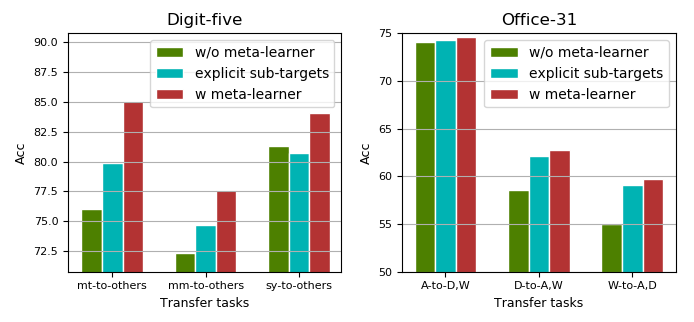}\vspace{-4pt}
	\caption{Ablation studies of our meta-learner across three transfer tasks in Digit-five (Backbone-2) and in Office-31 (AlexNet).} \label{DB}\vspace{-8pt}
\end{figure}

In \textbf{Digit-five} (Table.\ref{t1}), ANT frequently occurs in BTDA setups. In sv$\rightarrow$mm,mt,up,sy and sy$\rightarrow$mm,mt,up,sv, few DA algorithms are exempt for these performance degradation under backbone-1 (\emph{i.e.,} -26.8 for ADDA). This situation is ameliorated on backbone-2. However, all DA baselines suffer from RNT. Especially when an adaptation process starts from a simple source to complex targets, \emph{i.e.}, mt$\rightarrow$mm,sv,up,sy, the accuracy drop from MTDA to BTDA is significant. 
In \textbf{Office-31.} (Table.\ref{t2}), due to D and W sharing similar visual appearances, most of their categories are aligned well in the mixed target. Therefore ANT and RNT are suppressed in A$\rightarrow$ D,W. But in the other transfer tasks, ANT and RNT still haunt the performances of DA baselines. 
In \textbf{Office-Home} (Table.\ref{t3}), all DA baselines get rid of ANT, whereas they remain inferior to their performances in MTDA setup (suffer RNT). Observe that, deeper models, \emph{e.g.}, ResNet-50, may encourage the baselines to resist ANT. But the deeper models do not help reduce RNTs across DA algorithms. These evidences sufficiently verify the hardness of BTDA and answer the first question. 

Though BTDA is a challenging transfer setup, \textbf{AMEAN} presents as a ideal solver. As shown in Table.\ref{t1} -\ref{t3} , AMEAN achieved the state-of-the-art in $29$ out of $30$ BTDA transfer cases, and its average accuracy exceeds the second best by $1.0\sim7.7\%$. AMEAN almost achieve positive transfer in all transfer cases, and has reaped huge transfer gains in some of them, \emph{e.g.}, $+37.7\%$ in mt$\rightarrow$mm,sv,up,sy, $+21.5\%$ in A$\rightarrow$ D,W, $+31.3\%$ in Ar$\rightarrow$Cl,Pr,Rw, \hspace{-0.4em}\emph{etc}. \hspace{-.6em}More importantly, AMEAN obtains more impressive performances from deeper architectures, which demonstrates its superiority to address BTDA problem.

\vspace{-4pt}\subsection{Analysis.}\vspace{-6pt}

\textbf{Adaptation visualization.} For the BTDA transfer task \textbf{mt} $\rightarrow$ \textbf{mm},\textbf{sv},\textbf{up},\textbf{sy} in Digit-five, we visualize the classification activations from Source-only, RevGred, VADA and AMEAN in BTDA setup. As can be seen in Fig.\ref{vis} , Source-only barely captures any classification patterns. In a comparison, RevGred and VADA show better classifiable visualization patterns than Source-only's. But their activations remain pretty messy and most of them are misaligned in their classes. It demonstrates that BTDA is a very challenging scenario for existing DA algorithms. Finally, the activations from AMEAN show clear classification margins. It illustrates the superior transferability of our AMEAN.

\textbf{Ablation study.} The crucial component of our AMEAN is the meta-learner for auto-designing $V_{\rm mt}$. Hence our ablation focuses on this model-driven auto-learning technique. In particular, we evaluate three adaptation manners derived from our AMEAN: adaptation without meta-learner (w/o meta learner); adaptation without meta-learner but using explicit sub-target ($\{\mathcal{T}_j\}^k_{j=1}$) to guide the transfer in BTDA (explicit sub-targets); adaptation with meta-learner (w meta-learner, AMEAN). As illustrated in Fig.\ref{DB} , explicit sub-target information is not persistently helpful to BTDA. In \textbf{sy}-to-others, explicit sub-target information even draws back the source-to-target transfer gain. By contrast, meta-learner plays a key role to enhance the adaptation towards digit and real-world visual domains and obtain state-of-the-art in BTDA. Surprisingly, the dynamical meta-sub-targets even drive the adaptation model exceed those trained with the explicit sub-target domains. 

\begin{figure}[t]\centering
	\includegraphics[height=.9in,width=3.3in]{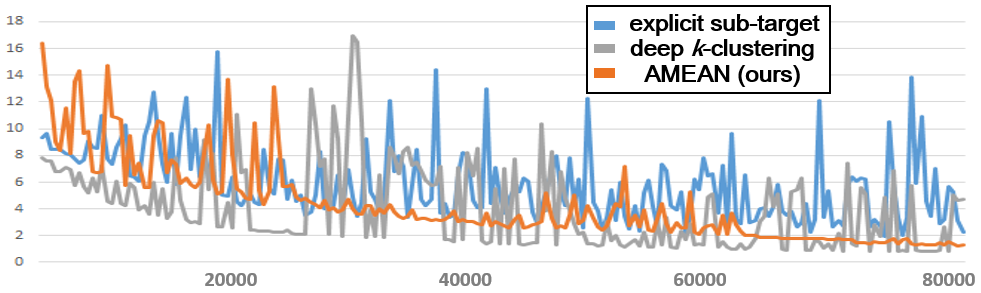}
	\vspace{-14pt}\caption{The ablation of $V_{\rm mt}$ based on mt$\rightarrow$mm,sv,up,sy in Digit-five, where $V_{\rm mt}$ is constructed in different manners over iterations. AMEAN performs a faster and more stable optimization process.
	}\label{loss}\vspace{-10pt}
\end{figure}

To further investigate the meta-adaptation dynamic provided by AMEAN, we ablate the multi-target DA by following different target separation strategies: 1). explicit-sub-target (EST); 2).static deep $k$-clustering ($k$-C); 3) AMEAN. Note that, 2) ablates the auto-loss-design dynamic in $V_{\rm mt}$, as it solely uses the initial clusters to divide the mixed target and keep a static $V_{\rm mt}$ along model training. The comparison of 2) and 3) helps to unveil whether our auto-loss strategy facilitates AMEAN. As shown in Table \ref{cl} , 1) and 2) disregard label information given by a source and their $V_{\rm mt}$ still suffer from a risk of class mismatching. Our auto-loss manner adaptively changes $V_{\rm mt}$ by receiving label information from the features previously learned by Eq.\ref{eq1.1} and thus, achieve better performance in BTDA. More importantly, it also encourages a fast and more stable adaptation during the minimax optimization process (see Fig \ref{loss}).    

\begin{table}[t]
	\center
	\caption{The ablation of $V_{\rm mt}$ constructed by Explicit Sub-Target (EST), deep $k$-Clustering ($k$-C) and AMEAN (ours).}
	\vspace{-10pt}\begin{footnotesize}
		\begin{tabular}{|l|cc|cc|ccr}
			&\hspace{-0.6em}mt$\rightarrow$mm,sv,up,sy	  	&\hspace{-0em}mm$\rightarrow$mt,sv,up,sy\hspace{-0.6em}	&D$\rightarrow$A,W	&W$\rightarrow$A,D						\\	
			\hline				
			EST&79.9&74.7&62.2&58.4 \\
			$k$-C&81.6&74.5&61.3&59.1 \\
			ours&\textbf{85.1}&\textbf{77.6}&\textbf{62.8}&\textbf{59.7} \\
		\end{tabular}
	\end{footnotesize}\label{cl}\vspace{-8pt}\end{table}

\vspace{-6pt}\section{Concluding Remarks}\vspace{-6pt}
In this paper, we concern a realistic adaptation scenario, where our target domain is comprised of 
multiple hidden sub-targets and learners could not identify which sub-target each unlabeled example comes from. This Blending-target domain adaptation (BTDA) conceives category mismatching risk and if we apply existing DA algorithm in BTDA, it will lead to negative transfer effect. To take on this uprising challenge, we propose Adversarial MEta-Adaptation Network (AMEAN). AMEAN starts from the popular adversarial adaptation methods, while newly employs a meta-learner network to dynamically devise a multi-target DA loss along with learning domain-invariant features. The AutoML merits are inherited by AMEAN to reduce the class mismatching in the mixed target domain. Our experiments focus on verifying the threat of BTDA and the efficacy of AMEAN in BTDA. Our evaluations show the significance of BTDA problem as well as the superiority of our AMEAN.   

\clearpage

\section{Appendix.A}

\subsection{Unsupervised Meta-learner}

Our meta-learner $U$ is trained as deep embedding clustering (\cite{xie2016unsupervised}) by receiving data and its feature-level feedbacks (the concatenation of the feature input to the discriminators and its classification result, for brevity, we mark $F(\boldsymbol{x}^{(t)})$ in our paper). It is very important to note that, distinguished from the original version solely using a DAE to initiate data embeddings, our meta-learner leverage the improved DEC \cite{guo2017improved} as our implementation, where the clustering embeddings are updated by reconstruction loss as well as the clustering objective \emph{w.r.t.} centroids $\{\mu_j\}^k_{j=1}$. Therefore, $U_1$, $U_2$ and $\{\mu_j\}^k_{j=1}$ are alternatively updated by
\begin{small}
	\begin{equation}\begin{aligned}
	U_2=&U_2 - \frac{\alpha}{m}\sum_{i=1}^{m}\frac{\partial L_{\rm rec}(\boldsymbol{x}_i; F) }{\partial U_2} \\
	U_1=&U_1 - \frac{\alpha}{m}\sum_{i=1}^{m}[\frac{\partial L_{\rm rec}(\boldsymbol{x}_i; F)-\sum_{j=1}^{k} p_{i,j}\log\frac{p_{i,j}}{q_{i,j}}}{\partial U_1}]\\
	\mu_j =& \mu_j - 2\frac{\alpha}{m}\sum_{i=1}^{m}\sum_{j=1}^{K}(1+||U_1(\boldsymbol{x}^{(t)}_i)-\mu_j||)^{-1}\\& \ \ \ \ \ \ \ \ \ \ \ \ \ \ \ \ \ \ \ \ \ \ \ \ \ \ \ \ \ \ \ \ \ \ \ \ \ \ \ \ \ (p_{ij}-q_{ij})(U_1(\boldsymbol{x}^{(t)}_i)-\mu_j) 
	\end{aligned}
	\end{equation}
\end{small}where $\alpha$ and $m$ denote the learning rate and mini-batch size for optimizing meta-learner. We set initial learning rate as $0.001$ and batch size is 256. The auto-encoder architecture implemented in our experiments has been shown in Table.\ref{dae}. 
\begin{table}
	\centering
	\caption{ The architecture of our unsupervsied meta-learner.  }
	\begin{small}
		\begin{tabular}{|l|c|c|c|c|c|cccr}
			\thickhline
			&\multirow{2}{1.5cm}{Input size} &\multirow{2}{1.5cm}{Output size} &\multirow{2}{1cm}{Activator}  \\
			
			& & &  \\ \hline
			\textbf{Encoder}: & & & \\
			En\_Fc\_1 &image size &500&ReLU\\
			En\_Fc\_2 &500&1000&ReLU\\
			En\_Fc\_3  &1000&$k$&ReLU\\
			\textbf{Decoder}: & & & \\
			De\_Fc\_1 &$k$&1000&ReLU\\
			De\_Fc\_2 &1000&1000&ReLU\\
			De\_Fc\_3 &1000&image-size&ReLU\\			
			\thickhline
		\end{tabular}
	\end{small}\label{dae}\end{table}
\subsection{Entropy Penalty}
In our implementation, we leverage a well-known cluster assumption \cite{grandvalet2005semi} to regulate the classifier $C$ learning with unlabeled target data. It can be interpreted as the minimization of the conditional entropy term with respect to the output of $C\big(F(\boldsymbol{x})\big)$
\begin{equation}
L_{\rm ent}(F,C)=-\mathbb{E}_{\boldsymbol{x}\sim\mathcal{T}} \ C\big(F(\boldsymbol{x})\big)^T\log \ C\big(F(\boldsymbol{x})\big)
\end{equation}. 
\begin{table*}
	\centering
	\caption{ Backbone-1 in digit-five experiment.  }
	\begin{small}
		\begin{tabular}{|l|c|c|c|c|c|cccr}
			\thickhline
			&\multirow{2}{1.5cm}{Kernel size} &\multirow{2}{2.5cm}{Output dimension} &\multirow{2}{1cm}{BN/IN}&\multirow{2}{1.5cm}{Activation} &\multirow{2}{1cm}{Dropout}  \\
			
			& & & && \\ \hline
			\textbf{Feature extractor}: & & & &&\\
			Conv1\_1	&5*5		&64*24*24		&BN	&ReLU&0  		 \\
			Maxpool     &2*2        &64*12*12       && &0.5 \\
			Conv1\_2	&5*5		&50*8*8		&BN	&ReLU&0  		 \\	
			Maxpool     &2*2        &50*4*4       && &0.5 \\	
			\textbf{Classifier}: & & & &&\\
			Fc\_1	&50*4*4		&100		&BN	&ReLU&0.5  		 \\
			Fc\_2	&100		&100		&BN	&ReLU&0  		 \\
			Fc\_3   &100&10&&Softmax&0\\			
			\textbf{Discriminator}: & & & &&\\
			Reversed gradient layer & & & &&\\
			Fc &50*4*4&100&BN&ReLU&0\\
			Fc ($D_{\rm st}$) &100&2&&Softmax&0\\
			Fc ($D_{\rm mt}$) &100&4&&Softmax&0\\			
			\thickhline
		\end{tabular}
	\end{small}\label{b1}\vspace{6pt}
	\caption{ Backbone-2 in digit-five experiment.  }
	\begin{small}
		\begin{tabular}{|l|c|c|c|c|c|cccr}
			\thickhline
			&\multirow{2}{1.5cm}{Kernel size} &\multirow{2}{2.5cm}{Output dimension} &\multirow{2}{1cm}{BN/IN}&\multirow{2}{1.5cm}{Activation} &\multirow{2}{1cm}{Dropout}  \\
			
			& & & && \\ \hline
			\textbf{Feature extractor}: & & & &&\\
			Conv1\_1	&3*3		&64*32*32		&IN/BN	&LeakyReLU(0.1)&0  		 \\
			
			Conv1\_2	&3*3		&64*32*32		&BN	&LeakyReLU(0.1)&0  		 \\
			
			Conv1\_3	&3*3		&64*32*32		&BN	&LeakyReLU(0.1)&0.5  		 \\
			
			Maxpool	&2*2		&64*16*16		&	&&  		 \\
			Conv2\_1	&3*3		&64*16*16		&BN	&LeakyReLU(0.1)&0  		 \\
			Conv2\_2	&3*3		&64*16*16		&BN	&LeakyReLU(0.1)&0  		 \\
			Conv2\_3	&3*3		&64*16*16		&BN	&LeakyReLU(0.1)&0.5  		 \\
			Maxpool	&2*2		&64*8*8		&	&&  		 \\			
			\textbf{Classifier}: & & & &&\\
			Conv2\_1	&3*3		&64*8*8		&BN	&LeakyReLU(0.1)&0  		 \\
			Conv2\_2	&3*3		&64*8*8		&BN	&LeakyReLU(0.1)&0  		 \\
			Conv2\_3	&3*3		&64*8*8		&BN	&LeakyReLU(0.1)&0  		 \\
			Averagepool	&		&64*1*1		&	&&  		 \\
			Fc &64*10&10&&Softmax&0\\			
			\textbf{Discriminator}: & & & &&\\
			Fc &64*8*8+10&100&&ReLU&0\\
			Fc ($D_{\rm st}$) &100*1&1&&Sigmoid&0\\
			Fc ($D_{\rm mt}$) &100*4&4&&Softmax&0\\			
			\thickhline
		\end{tabular}
	\end{small}\label{b2}	
\end{table*}
The objective forces the classification to be confident on the unlabeled target example, which drives the classifier's decision boundaries away from the target unlabeled examples. It has been applied in wide range of domain adaptation researches \cite{sankaranarayanan2017generate} \cite{long2016unsupervised} \cite{gebru2017fine}. However, while using available data to empirically estimate the expected loss, \cite{grandvalet2005semi} demonstrates that such approximation provably breaks down if $C\big(F(\cdot)\big)$ does not satisfy local Lipschitz condition. Specifically, the classifier without local Lipschitz constraint can abruptly changes its prediction, which allows placement of the classifier decision boundaries close to target training examples while the empirical conditional entropy is still minimized. To prevent this issue, we follow the technique in \cite{shu2018dirt} where virtual adversarial perturbation term \cite{miyato2018virtual} is incorporated to regulate the classifier and feature extractor:
\begin{equation}\begin{aligned}
L_{\rm vir}(F,&C)=  \ \ \ \ \ \ \ \ \ \ \\\underset{\boldsymbol{x}^{(s)}\sim\mathcal{S}}{\mathbb{E}} \big[&\underset{||\mathbf{r}||\leq \epsilon}{\max} \mathbf{D}_{\rm KL} (C\big(F(\boldsymbol{x}^{(s)})\big)||C\big(F(\boldsymbol{x}^{(s)}+\mathbf{r})\big)) \big]\\+&\rho\mathbb{E}_{\boldsymbol{x}^{(t)}\sim\mathcal{T}} \big[\underset{||\mathbf{r}||\leq \epsilon}{\max} \mathbf{D}_{\rm KL} (C\big(F(\boldsymbol{x}^{(t)})\big)||C\big(F(\boldsymbol{x}^{(t)}+\mathbf{r})\big)) \big]
\end{aligned}
\end{equation}where $\mathbf{D}_{\rm KL}$ indicates KL divergence. $\mathbf{r}$ indicates the virtual adversarial perturbation upper bounded by a magnitude $\epsilon>0$ on source and target images ($\boldsymbol{x}^{(s)}\in \mathcal{S}$ and $\boldsymbol{x}^{(t)}\in \mathcal{T}$ ), which are obtained by maximizing the classification differences between $C\big(F(\boldsymbol{x}^{(s)})\big)$ and $C\big(F(\boldsymbol{x}^{(s)}+\mathbf{r})\big)$. This restrictions are simultaneously proposed on source and target and $\rho$ is the balance factor between them. In this way, the collaborative meta-adversarial adaptation objectives (Eq.8-10 in our paper) are reformulated as:\begin{table*}
	\centering
	\caption{ The hyper-parameters setting in our experiment.  }\label{d}
	\begin{small}
		\begin{tabular}{|l|c|c|c|c|ccccr}
			\thickhline
			&\multicolumn{2}{c|}{Office-31, Office-HOME}&\multicolumn{2}{c|}{Digit-five}\\\hline
			&\multirow{2}{*}{AlextNet} &\multirow{2}{*}{ResNet-50} &\multirow{2}{*}{Backbone-1} &\multirow{2}{*}{Backbone-2}  \\
			
			& & & & \\ \hline
			\multirow{2}{2cm}{mini-batch size}	&\multirow{2}{*}{32}&\multirow{2}{*}{32}		&\multirow{2}{*}{128}		&\multirow{2}{*}{100}	  		 		\\& & &&  \\
			\hline
			\multirow{2}{*}{$\lambda$}  &\multirow{2}{*}{0.1}	&\multirow{2}{*}{1}&\multirow{2}{*}{1}&\multirow{2}{*}{1 (update $D_{\rm st}$) / 0.01 (update $F$)}	  	 \\& & &&	\\
			\hline												
			\multirow{2}{1cm}{$\gamma$	}&\multirow{2}{*}{0.01}
			&\multirow{2}{*}{$\frac{\rm iter}{\rm max_iter}$}				&\multirow{2}{*}{0.1}		&\multirow{2}{*}{$\frac{\rm iter}{\rm max_iter}$} \\
			& & &&	\\\hline												
			\multirow{2}{1cm}{$\beta$}
			&\multirow{2}{*}{0.01}&\multirow{2}{*}{0.1}				&\multirow{2}{*}{0}		&\multirow{2}{*}{0.01} \\
			& && &	\\\hline
			\multirow{2}{1cm}{$\rho$}
			&\multirow{2}{*}{0.01}&\multirow{2}{*}{0}				&\multirow{2}{*}{0}		&\multirow{2}{*}{0.01} \\
			& && &	\\\hline
			\multirow{2}{1cm}{$M$	}
			&\multirow{2}{*}{2000}&\multirow{2}{*}{2000}				&\multirow{2}{*}{20000}		&\multirow{2}{*}{10000} \\
			&& & &	\\\hline
			\multirow{2}{*}{image size} &\multirow{2}{*}{227$\times$227}&\multirow{2}{*}{227$\times$227}				&\multirow{2}{*}{28$\times$28}		&\multirow{2}{*}{28$\times$28} \\
			&&&&\\			
			\thickhline
		\end{tabular}
	\end{small}	
\end{table*}
\begin{equation}
\begin{aligned}
\underset{D_{\rm st}, D_{\rm mt}}{\max} \underset{F, C}{\min} \ \ &V_{\rm joint}(D_{\rm st},D_{\rm mt},F, C) \\&= V_{\rm st}(F, D_{\rm st}, C) +\gamma V_{\rm mt}(F, D_{\rm mt}) \\
& \ \ \ \ \ \ \ \ \ \ \ \ +\beta L_{\rm ent}(F,C)+L_{\rm vir}(F,C)\label{eq.joint}
\end{aligned}
\end{equation}and
\begin{equation}
\begin{aligned}
\underset{D_{\rm st}, D_{\rm mt}}{\max} \ V_{\rm alter}(D_{\rm st},D_{\rm mt}) = V_{\rm st}(F, D_{\rm st}, C)+ V_{\rm mt}(F, D_{\rm mt})\label{eq.alt1}
\end{aligned}
\end{equation}
\begin{equation}
\begin{aligned}
\underset{F, C}{\min} \ V_{\rm alter}(F, C) = &V_{\rm st}(F, D_{\rm st}, C)+\gamma \widetilde{V}_{\rm mt}(F, D_{\rm mt})\\
& \ \ \ \ \ \ \ \ \ \ \ \ +\beta L_{\rm ent}(F,C)+L_{\rm vir}(F,C)\label{eq.alt2}
\end{aligned}
\end{equation}. We provide the ablation study of the entropy penalty in Table \ref{et} .
\begin{table}[h]
	\caption{Some ablation of entropy term.}\begin{footnotesize}
		\begin{tabular}{|l|cc|cc|ccr}
			\hline
			&mt$\rightarrow$mm,sv,up,sy	  	&mm$\rightarrow$mt,sv,up,sy	&D$\rightarrow$A,W	&W$\rightarrow$A,D						\\					
			w&\textbf{85.1}&\textbf{77.6}&\textbf{62.8}&\textbf{59.7} \\
			w/o	&83.7&76.9&62.6&59.2 \\\hline
		\end{tabular}
	\end{footnotesize}\label{et}
\end{table}

\subsection{The Selection of $k$}
In AMEAN, $k$ denotes the number of sub-target domains and is pre-given. Table \ref{kstudy} demonstrates that choosing $k$ as the number of sub-targets leads to the superior performance of AMEAN. However, whether AMEAN would achieve the better performance if $k$ is adaptively determined, remains an open and interesting question. We would like to investigate this topic in the future.

\begin{table}[h]\centering
	\caption{\textbf{Acc} is the average accuracy over five sub-transfer tasks in Digit-five. The number of sub-targets in Digit-five is $4$. }\label{kstudy}
	\begin{footnotesize}
		\begin{tabular}{|l|ccccccc|cr}
			\hline
			\emph{k} &2 &3&4&5&6&7&8		\\					
			\textbf{Acc} &79.3 &83.2 &\textbf{83.7} &82.1 &83.2 &82.0 &78.6\\\hline 
		\end{tabular}
	\end{footnotesize}\vspace{-16pt}
\end{table}

\subsection{Architectures}\vspace{-5pt}
The architectures for digit recognition in Digit-five have been illustrated in Table.\ref{b1} , \ref{b2} . The first backbone is based on LeNet and the second is derived from \cite{shu2018dirt} for comparing their state-of-the-art models VADA and DIRT-T. The architectures for object recognition in Office-31 and Office-Home are based on AlexNet and ResNet-50, which are consistent with the previous studies \cite{long2015learning} \cite{long2016deep} \cite{long2016unsupervised} \cite{ganin2014unsupervised} . 
\subsection{Training Details}
\vspace{-5pt}We evenly separate the proportion of the source and target examples in each mini-batch. Concretely, we promise that a half of examples in a mini-batch are drawn from $\mathcal{S}$ and the rest belong to the mixed target domain training set $\mathcal{T}^{\rm train}$: In \emph{digit-five}, we randomly drew target examples from the mixed target set $\mathcal{T}^{\rm train}$ to construct our mini-batches; In \emph{Office-31} and \emph{Office-Home}, we promise the number of target examples from different meta-sub-target are the same by repeat sampling. 

In the Digit-five experiment, we add a confusion loss \cite{hoffman2017simultaneous} \emph{w.r.t.} $\mathcal{S}$ to train the backbone-2. It stabilizes the alternating adaptation since the mixed target in Digit-five is more diverse than the other benchmarks' and the alternating learning manner is quite instable in these scenarios. The implementation can be found in our code.

The hyper-parameters are shown in Table \ref{d} .

\section{Appendix.B}
\subsection{Evalutation Metrics for BTDA}
We elaborate how to calculate ANT and RNT in our experiment in details:
\begin{equation}
ANT = \max\{0, Acc_{\rm BTDA}-Acc_{\rm Source-only}\}\label{ant}
\end{equation}where $Acc_{\rm Source-only}$ denotes the classification accuracy about the model trained on the source labeled dataset $\mathcal{S}$ and tested on the mixed target set $\mathcal{T}^{\rm test}=\underset{j=1}{\overset{k}{\cup}}\mathcal{T}^{\rm test}_j$; $Acc_{\rm BTDA}$ denotes the multi-target-weighted classification accuracy of the evaluated DA model under BTDA setup:
\begin{equation}
Acc_{\rm BTDA}=\sum_{j=1}^{k}\alpha_jAcc_{\rm BTDA}^{(j)}\label{acc}
\end{equation}where $Acc_{\rm BTDA}^{(j)}$ denotes the DA model classification accuracy on the $j^{th}$ sub-target domain test set $\mathcal{T}_j^{\rm test}$ when the evaluated DA models (\emph{e.g.}, JAN, DAN, AMEAN, \emph{etc}) is trained with the source labeled set $\mathcal{S}$ and the mixed target unlabeled set $\mathcal{T}^{\rm train}=\underset{j=1}{\overset{k}{\cup}}\mathcal{T}^{\rm train}_j$ (BTDA setup). $\{\alpha_j\}^k_{j=1}$ denotes the proportion of the multi-target mixture. It is derived from the domain-set proportion in benchmarks, which are valued by $\{0.236,0.236,0.236,0.236,0.056\}$, $\{0.686, 0.121, 0.193\}$ and $\{0.155, 0.280, 0.285, 0.280\}$ in Digit-five, Office-31 and Office-Home. When we draw the subset of domains to construct the mixed target, $\{\alpha_j\}^k_{j=1}$ is obtained by normalizing these corresponding benchmark-specific domain-set proportion \footnote{The numbers are based on the hidden sub-target test set proportions in a mixed target} . 

\textbf{In reality, we can obtain $Acc_{\rm BTDA}$ by directly evaluating the DA models on the mixed test set $\mathcal{T}_j^{\rm test}$, which leads to the same results in (\ref{acc})}. 


Based on (\ref{acc}), we also define the RNT metric 
\begin{equation}
RNT = Acc_{\rm BTDA}-\sum_{j=1}^{k}\alpha_jAcc_{j}\label{rnt}
\end{equation}where $Acc_j$ denotes the $j^{th}$-target test classification accuracy with respect to a single-target DA classifier trained on the source labeled set $\mathcal{S}$ and the $j^{th}$ sub-target unlabeled set $\mathcal{T}^{\rm train}_j$. Note that,

\begin{itemize}
	\item $Acc_{j}$ is derived from a DA model trained with datasets $\mathcal{S}$ and $\mathcal{T}^{\rm train}_j$. It means that $Acc_{i}$, $Acc_{j}$ ($i\neq j$) are derived from different DA models, which employ the same DA algorithms yet are trained on $\mathcal{T}_i^{\rm train}$, $\mathcal{T}_j^{\rm train}$ and tested on $\mathcal{T}_i^{\rm test}$, $\mathcal{T}_j^{\rm test}$, respectively. 
	
	\vspace{-10pt}\item $Acc^{(j)}_{\rm BTDA}$ \hspace{-0.1em}is \hspace{-0.1em}\hspace{-0.1em} derived \hspace{-0.1em}from \hspace{-0.1em}a \hspace{-0.1em}DA \hspace{-0.1em}model\hspace{-0.1em} trained with $\mathcal{S}$ and $\mathcal{T}^{\rm train}=\underset{j=1}{\overset{k}{\cup}}\mathcal{T}^{\rm train}_j$. Hence $Acc^{(i)}_{\rm BTDA}$, $Acc^{(j)}_{\rm BTDA}$ ($i\neq j$) are derived from the same DA model, which employ the same DA algorithms and is trained on ($\mathcal{S}\cup\mathcal{T}^{\rm train}$) and then, tested on $\mathcal{T}_i^{\rm test}$ and $\mathcal{T}_j^{\rm test}$ to induce $Acc^{(i)}_{\rm BTDA}$, $Acc^{(j)}_{\rm BTDA}$.\vspace{-10pt}
\end{itemize}. 

\begin{table*}[h]
	\centering
	\caption{ The equal-weight (EW) Classification accuracy (ACC \%), \emph{absolute negative transfer} (ANT\%) and \emph{relative negative transfer} (RNT\%) on Digit-five in BTDA setup. \textbf{{\color{blue}BLUE}}, \textbf{{\color{red}RED}} indicate ANT and RNT, respectively. More viewed in (\ref{ew}). }\label{t1}
	\begin{scriptsize}
		\begin{tabular}{|l|cc|cc|cc|cc|cc|cc|r}\thickhline
			Models &\multicolumn{2}{c|}{mt$\rightarrow$mm,sv,up,sy} &\multicolumn{2}{c|}{mm$\rightarrow$mt,sv,up,sy} &\multicolumn{2}{c|}{sv$\rightarrow$mm,mt,up,sy} &\multicolumn{2}{c|}{sy$\rightarrow$mm,mt,sv,up} &\multicolumn{2}{c|}{up$\rightarrow$mm,mt,sv,sy} &\multicolumn{2}{c|}{Avg}\\
			&ACC$^{\rm ANT}$		&RNT	 
			&ACC$^{\rm ANT}$		&RNT	 
			&ACC$^{\rm ANT}$		&RNT	 
			&ACC$^{\rm ANT}$		&RNT	 
			&ACC$^{\rm ANT}$		&RNT	 
			&ACC$^{\rm ANT}$		&RNT	 
			\\		\hline
			\textbf{Backbone-1:}	&	&	&	&	& &	 &  &	 &  & &  &	\\
			Source only				&36.6 &0	&57.3 &0  	&67.1 &0	 &74.9 &0 &36.9 &0 &54.6 &0 		\\
			ADDA		&52.5 &{\color{red}-7.4}	&{58.9}&{\color{red}-1.2}	&{\color{blue}46.4$^{(-20.7)}$}&{\color{red}-16.0}	 &{\color{blue}67.0$^{(-7.9)}$}&{\color{red}-7.0} &{\color{blue}34.8$^{(-2.1)}$}&{\color{red}-13.3} &{\color{blue}51.9$^{(-2.7)}$}&{\color{red}-9.0}		\\
			DAN				&38.8&{\color{red}-8.6}	 &{\color{blue}53.5$^{(-3.8)}$}&{\color{red}-4.5}&{\color{blue}55.1$^{(-12.0)}$}&-{\color{red}3.0}&{\color{blue}65.8$^{(-9.1)}$}&{\color{red}-2.8} &{\color{blue}27.0$^{(-9.9)}$}&{\color{red}-11.0} 	&{\color{blue}48.0$^{(-6.6)}$}&{\color{red}-6.0}		\\
			GTA				&51.4&{\color{red}-9.0}	&{\color{blue}54.2$^{(-3.1)}$}&{\color{red}-2.1}	  	&{\color{blue}59.8$^{(-7.3)}$}&{\color{red}-3.6} &\textbf{76.2(+1.3)}&{\color{red}-0.6}&41.3&{\color{red}-2.0}&56.6&{\color{red}-3.6}		\\
			RevGrad		&60.2&{\color{red}-6.2}&66.0&{\color{red}-4.6}&{\color{blue}64.7$^{(-2.3)}$}&{\color{red}-6.0}&{\color{blue}69.2$^{(-5.7)}$}&{\color{red}-7.1}	  	&44.3&{\color{red}-6.3} &60.9&{\color{red}-6.0} 		\\
			AMEAN   &\textbf{61.2 (+1.0)}&-	&\textbf{66.9 (+0.9)}&-	&\textbf{67.2 (+0.1)}	&-  	&{\color{blue}73.3$^{(-1.6)}$}	&-	 &\textbf{47.5 (+3.2)}  &- &\textbf{63.2 (+2.3)} &-\\
			\hline
			\textbf{Backbone-2:}	&	&	&	&	& &	 &  &	 &  & &  &	\\
			Source only				&55.8 &0	&55.2  &0  	&74.3 &0 &76.4 &0 &50.6 &0  &62.5 &0 		\\
			VADA				&79.4&{\color{red}-4.9}	  	&72.5&{\color{red}-3.1} &76.4&{\color{red}-2.2}&82.8&{\color{red}-3.8}&56.4&{\color{red}-8.7} &73.5&{\color{red}-4.5} 		\\
			DIRT-T				&77.5&{\color{red}-6.5} &76.8&{\color{red}-4.4}&\textbf{79.7 (+1.8)}&{\color{red}-4.9}&80.9&{\color{red}-3.9}&47.0&{\color{red}-7.5}&72.4&{\color{red}-5.5} 		\\
			AMEAN  &\textbf{86.9 (+7.5)}&-	&\textbf{78.5 (+1.7)} &-	&77.9	&-  	&\textbf{85.6 (+2.8)}	&-	 &\textbf{75.5 (+19.1)}  &- &\textbf{80.9 (+7.4)} &-\\
			\hline
		\end{tabular}
	\end{scriptsize}
\end{table*}

\begin{table*}[h]
	\centering
	\caption{The equal-weight (EW) Classification accuracy (ACC \%), \emph{absolute negative transfer} (ANT\%) and \emph{relative negative transfer} (RNT\%) on Office31 in BTDA setup. \textbf{{\color{blue}BLUE}}, \textbf{{\color{red}RED}} indicate ANT and RNT, respectively. More viewed in (\ref{ew}).}\label{t2}
	\begin{footnotesize}
		\begin{tabular}{|lc|cc|cc|cc|cc|r}\thickhline
			\multirow{2}{*}{Backbones}&\multirow{2}{*}{Models} &\multicolumn{2}{c|}{A$\rightarrow$D,W} &\multicolumn{2}{c|}{D$\rightarrow$A,W} &\multicolumn{2}{c|}{W$\rightarrow$A,D} &\multicolumn{2}{c|}{Avg}\\
			&&ACC$^{\rm ANT}$		&RNT	&ACC$^{\rm ANT}$		&RNT	&ACC$^{\rm ANT}$		&RNT	&ACC$^{\rm ANT}$		&RNT			\\\hline
			\multirow{6}{*}{\color{red}AlexNet}&Source only	 & 62.7 &0		& 73.3 &0		& 74.4 &0	& 70.1 &0	\\	
			&DAN				&68.2 & 0.0	&{\color{blue}71.4$^{(-1.9)}$}&{\color{red}-4.0}		&{\color{blue}73.2$^{(-1.2)}$}&{\color{red}-3.3}&70.9&{\color{red}-2.4}		\\
			&RTN				&70.7&{\color{red}-1.7}	&{\color{blue}69.8$^{(-3.5)}$}&{\color{red}-4.1}		&{\color{blue}71.5$^{(-2.9)}$}&{\color{red}-3.9} &70.7&{\color{red}-3.2}		\\
			&JAN				&73.5&{\color{red}-0.1}		&73.6&{\color{red}-4.1}		 &75.0&{\color{red}-2.5} &74.0&{\color{red}-2.2}		\\
			&RevGrad				&74.1&1.0	 &{\color{blue}72.1$^{(-1.2)}$}&{\color{red}-2.8}  	&{\color{blue}73.4$^{(-1.0)}$}&{\color{red}-1.8}&73.2&{\color{red}-1.2}		\\
			&AMEAN (ours)  	&\textbf{74.9 (+0.8)}	&-  	&\textbf{74.9 (+1.3)}	&-	 &\textbf{76.2 (+1.2)}  &- &\textbf{75.3 (+1.3)} &-\\
			\hline
		\end{tabular}
	\end{footnotesize}
	\caption{The equal-weight (EW) Classification accuracy (ACC \%), \emph{absolute negative transfer} (ANT\%) and \emph{relative negative transfer} (RNT\%) on Office31 in BTDA setup. \textbf{{\color{blue}BLUE}}, \textbf{{\color{red}RED}} indicate ANT and RNT, respectively. More viewed in (\ref{ew}).}\label{t4}
	\begin{footnotesize}
		\begin{tabular}{|lc|cc|cc|cc|cc|r}\thickhline
			\multirow{2}{*}{Backbones}&\multirow{2}{*}{Models} &\multicolumn{2}{c|}{A$\rightarrow$D,W} &\multicolumn{2}{c|}{D$\rightarrow$A,W} &\multicolumn{2}{c|}{W$\rightarrow$A,D} &\multicolumn{2}{c|}{Avg}\\
			&&ACC$^{\rm ANT}$		&RNT	&ACC$^{\rm ANT}$		&RNT	&ACC$^{\rm ANT}$		&RNT	&ACC$^{\rm ANT}$		&RNT			\\\hline
			\multirow{6}{*}{\color{red}ResNet-50}&Source only	 &68.7 &0		&79.6 &0		&80.0 &0	&76.1 &0	\\	
			&DAN				&77.9&{\color{red}-2.0}	&{\color{blue}75.0$^{(-4.6)}$}&{\color{red}-5.0}	&80.0&{\color{red}-1.3}&77.6&{\color{red}-3.0}		\\
			&RTN				&84.1&+2.9		&{\color{blue}77.2$^{(-2.4)}$}&{\color{red}-4.4}	 &{\color{blue}79.0$^{(-1.0)}$}&{\color{red}-3.3} &80.1&{\color{red}-1.6}		\\
			&JAN				&84.6&{\color{red}-0.8}		&82.7&{\color{red}-0.6}		 &83.4&{\color{red}-1.8} &83.6&{\color{red}-1.0}		\\
			&RevGrad				&79.0&{\color{red}-2.3}	 &81.4&{\color{red}-1.5}	  	&82.3&{\color{red}-1.3}&80.9&{\color{red}-1.7}		\\
			&AMEAN (ours)  	&\textbf{89.8 (+5.2)}	&-  	&\textbf{84.6 (+1.9)}	&-	 &\textbf{84.3 (+0.9)}  &- &\textbf{86.2 (+2.6)} &-\\
			\hline
		\end{tabular}
	\end{footnotesize}
	\caption{The equal-weight (EW) Classification accuracy (ACC \%), \emph{absolute negative transfer} (ANT\%) and \emph{relative negative transfer} (RNT\%) on OfficeHome in BTDA setup. \textbf{{\color{blue}BLUE}}, \textbf{{\color{red}RED}} indicate ANT and RNT, respectively. More viewed in (\ref{ew}).}\label{t3}\vspace{-0pt}
	\begin{footnotesize}
		\begin{tabular}{|lc|cc|cc|cc|cc|cc|r}\thickhline
			\multirow{2}{*}{Backbones}&\multirow{2}{*}{Models} &\multicolumn{2}{c|}{Ar$\rightarrow$Cl,Pr,Rw} &\multicolumn{2}{c|}{Cl$\rightarrow$Ar,Pr,Rw} &\multicolumn{2}{c|}{Pr$\rightarrow$Ar,Cl,Rw} &\multicolumn{2}{c|}{Rw$\rightarrow$Ar,Cl,Pr} &\multicolumn{2}{c|}{Avg}\\
			&&ACC$^{\rm ANT}$		&RNT	&ACC$^{\rm ANT}$		&RNT	&ACC$^{\rm ANT}$		&RNT	&ACC$^{\rm ANT}$		&RNT	&ACC$^{\rm ANT}$		&RNT		\\\hline
			\multirow{6}{*}{\color{red}AlexNet}&Source only				&33.4 &0 &35.3 &0 &30.6 &0&37.9 &0&34.3 &0 		\\
			&DAN				&39.7&{\color{red}-3.6}	 &41.6&{\color{red}-2.8}	&37.8&{\color{red}-3.1}&46.8&{\color{red}-2.4}&41.5&{\color{red}-3.0}	\\
			&RTN				&42.8&{\color{red}-2.0} &43.4&{\color{red}-2.4} &39.1&{\color{red}-2.1}	  			 &48.8&{\color{red}-2.5}  &43.5&{\color{red}-2.2}		\\
			&JAN				&43.5&{\color{red}-2.9} &44.6&{\color{red}-3.5} &39.4&{\color{red}-5.2}	  			 &48.5&{\color{red}-5.2}  &44.0&{\color{red}-4.2}		\\
			&RevGrad				&42.2&{\color{red}-3.3}& 43.8&{\color{red}-3.5}		&39.9&{\color{red}-3.6}	  	&47.7&{\color{red}-5.0} &43.4&{\color{red}-3.9}		\\
			&AMEAN (ours)  	&\textbf{44.6 (+1.1)}	&-  &\textbf{45.6 (+1.0)}	&-	 &\textbf{41.4 (+1.5)}  &- &\textbf{49.3 (+0.5)} & &\textbf{45.2 (+1.2)} &-\\
			\hline
		\end{tabular}
	\end{footnotesize}
	\center
	\caption{The equal-weight (EW) Classification accuracy (ACC \%), \emph{absolute negative transfer} (ANT\%) and \emph{relative negative transfer} (RNT\%) on OfficeHome in BTDA setup. \textbf{{\color{blue}BLUE}}, \textbf{{\color{red}RED}} indicate ANT and RNT, respectively. More viewed in (\ref{ew}).}\label{t5}\vspace{-0pt}
	\begin{footnotesize}
		\begin{tabular}{|lc|cc|cc|cc|cc|cc|r}\thickhline
			\multirow{2}{*}{Backbones}&\multirow{2}{*}{Models} &\multicolumn{2}{c|}{Ar$\rightarrow$Cl,Pr,Rw} &\multicolumn{2}{c|}{Cl$\rightarrow$Ar,Pr,Rw} &\multicolumn{2}{c|}{Pr$\rightarrow$Ar,Cl,Rw} &\multicolumn{2}{c|}{Rw$\rightarrow$Ar,Cl,Pr} &\multicolumn{2}{c|}{Avg}\\
			&&ACC$^{\rm ANT}$		&RNT	&ACC$^{\rm ANT}$		&RNT	&ACC$^{\rm ANT}$		&RNT	&ACC$^{\rm ANT}$		&RNT	&ACC$^{\rm ANT}$		&RNT		\\\hline
			\multirow{6}{*}{\color{red}ResNet-50}&Source only				&47.6 &0 &41.8 &0 &43.4 &0&51.7 &0&46.1 &0 		\\
			&DAN				&55.6&{\color{red}-0.5}	 &55.1&+0.9	&47.8&{\color{red}-4.0}&56.6&{\color{red}-6.3}&53.8&{\color{red}-2.5}	\\
			&RTN				&53.9&{\color{red}-1.8} &55.4&{\color{red}-0.7} &47.2&{\color{red}-3.3}	  			 &51.8&{\color{red}-3.0}  &52.1&{\color{red}-2.2}		\\
			&JAN				&58.3&{\color{red}-0.4} &59.2&+2.1 &51.9&{\color{red}-1.2}	  			 &57.8&{\color{red}-6.1}  &56.8&{\color{red}-1.5}		\\
			&RevGrad				&58.4&{\color{red}-3.0}& 57.0&{\color{red}-2.2}		&52.2&{\color{red}-4.6}	  	&62.0&{\color{red}-3.2} &57.4&{\color{red}-3.2}		\\
			&AMEAN (ours)  	&\textbf{64.3 (+5.9)}	&-  &\textbf{64.2 (+5.0)}	&-	 &\textbf{59.0 (+6.8)}  &- &\textbf{66.4 (+4.4)} & &\textbf{63.5 (+6.1)} &-\\
			\hline
		\end{tabular}
	\end{footnotesize}
\end{table*}


\textbf{Equal-weight ANT, RNT.} It worth noting that, though ANT/RNT in (\ref{ant}),(\ref{rnt}) are able to reflect BTDA models' performances on a mixed target domain set, it is not enough to demonstrate the comprehensive performances of the models over multi-sub-target domains, since it does not equally weight hidden sub-target domains. More specifically, imagine that we have a small set of target images belonging to a hidden sub-target, which the model performs poorly on. Then the RNT metric would shield the model's incapacity on that domain. 

In order to thoroughly reflect the capacities of evaluated models, we additionally report the results when the proportion $\{\alpha_j\}^k_{j=1}$ is equally set. In particular, we tend to consider the equal-weight classification accuracy ($Acc^{\rm (EW)}_{\rm BTDA}$), and quantify the corresponding negative transfer $ANT_{EW}$ and $RNT_{EW}$ in this setup:
\begin{equation}\begin{aligned}\label{ew}
Acc^{\rm (EW)}_{\rm BTDA}&=\frac{1}{k}\sum_{j=1}^{k}Acc_{\rm BTDA}^{(j)}\\
ANT_{EW} &= \max\{0, Acc^{\rm (EW)}_{\rm BTDA}-Acc^{\rm (EW)}_{\rm Source-only}\}\\
RNT_{EW} &= Acc^{\rm (EW)}_{\rm BTDA}-\frac{1}{k}\sum_{j=1}^{k}Acc_{j}
\end{aligned}
\end{equation}. The metrics developed from (\ref{ant} \ref{acc} \ref{rnt}) could be viewed as the complementary of what we report in the paper. 

\subsection{Evaluated Baselines in BTDA setup.} Beyond our AMEAN model, we also reported the BTDA performances from state-of-the-art DA baselines in Digit-five, Office-31, Office-Home. The baselines include Deep Adaptation Network (\textbf{DAN}) \cite{long2015learning}, Residual Transfer Network (\textbf{RTN}) \cite{long2016unsupervised}, Joint Adaptation Network (\textbf{JAN}) \cite{Long2017Deep}, Generate To Adapt (\textbf{GTA}) \cite{sankaranarayanan2017generate}, Adversarial Discriminative Domain Adaptation (\textbf{ADDA}) \cite{tzeng2017adversarial}, Reverse Gradient (\textbf{RevGrad}) \cite{ganin2014unsupervised} \cite{Ganin2017Domain}, Virtual Adversarial Domain Adaptation (\textbf{VADA}) \cite{shu2018dirt} and its variant \textbf{DIRT-T} \cite{shu2018dirt}. 

In the \emph{Digit-five} experiment, DAN, ADDA, GTA, ReGrad are all derived from their official codes. To promise a fair comparison, we standardize the backbones by LeNet to report $Acc_{\rm BTDA}$, $Acc^{\rm (EW)}_{\rm BTDA}$ and the negative transfer effects. VADA and DIRT-T are evaluated by their official codes to provide the results. Their model architectures are consistent with our backbone-2.

In the \emph{Office-31} and \emph{Office-Home} experiments, we employ the official codes of DAN, RTN, JAN, ReGrad to report $Acc_{\rm BTDA}$, $Acc^{\rm (EW)}_{\rm BTDA}$ in the \emph{Office-31} and \emph{Office-Home} experiments. 

The codes of all evaluated baselines can be found in their literatures. For a fair comparison, $Acc_{j}$ mainly originates from the reported results in their papers.


\subsection{BTDA experiments by equal-weight evaluation metrics}
The equal-weight versions of the classification accuracy ($Acc^{\rm (EW)}_{\rm BTDA}$), absolute negative transfer ($ANT_{EW}$) and relative negative transfer $RNT_{EW}$ over all the evaluated baselines in Digit-five, Office-31 and Office-Home are reported in Table \ref{t1}- \ref{t5}. 

{\small
	\bibliographystyle{ieee}
	\bibliography{reference}
}

\end{document}